\def\year{2020}\relax
\documentclass[letterpaper]{article} 
\usepackage{aaai20}  
\usepackage{times}  
\usepackage{helvet} 
\usepackage{courier}  
\usepackage[hyphens]{url}  
\usepackage{graphicx} 
\def\UrlFont{\rm}  
\usepackage{graphicx}  
\usepackage{amsmath}
\usepackage{amssymb}
\usepackage{multirow}
\usepackage{booktabs}
\usepackage{tabularx}
\usepackage{titling}

\newcommand{\citet}[1]
{\citeauthor{#1}~\shortcite{#1}}
\newcommand{\citep}{\cite}

\title{\textbf{Learning End-To-End Scene Flow by Distilling Single Tasks Knowledge}}
\author{\textbf{Filippo Aleotti, Matteo Poggi, Fabio Tosi, Stefano Mattoccia}\\
Department of Computer Science and Engineering (DISI) \\
University of Bologna, Italy}
\date{}

\begin{document}

\maketitle
\insert\footins{\noindent\footnotesize Copyright \copyright{ }2020,
Association for the Advancement of Artificial Intelligence (www.aaai.org).
All rights reserved.}

\begin{abstract}
Scene flow is a challenging task aimed at jointly estimating the 3D structure and motion of the sensed environment. Although deep learning solutions achieve outstanding performance in terms of accuracy, these approaches divide the whole problem into standalone tasks (stereo and optical flow) addressing them with independent networks. Such a strategy dramatically increases the complexity of the training procedure and requires power-hungry GPUs to infer scene flow barely at 1 FPS.
Conversely, we propose DWARF, a novel and lightweight architecture able to infer full scene flow jointly reasoning about depth and optical flow easily and elegantly trainable end-to-end from scratch.
Moreover, since ground truth images for full scene flow are scarce, we propose to leverage on the knowledge learned by networks specialized in stereo or flow, for which much more data are available, to distill proxy annotations.
Exhaustive experiments show that i) DWARF runs at about 10 FPS on a single high-end GPU and about 1 FPS on NVIDIA Jetson TX2 embedded at KITTI resolution, with moderate drop in accuracy compared to $10\times$ deeper models, ii) learning from many distilled samples is more effective than from the few, annotated ones available.
Code available at: \UrlFont{\textbf{https://github.com/FilippoAleotti/Dwarf-Tensorflow}}
   
\end{abstract}

\section{Introduction}
The term \emph{Scene Flow} refers to the three-dimensional dense motion field of a scene \cite{sceneflow} and enables to effectively model both 3D structure and movements of the sensed environment, crucial for a plethora of high-level tasks such as augmented reality, 3D mapping and autonomous driving. 
Dense scene flow inference requires the estimation of two crucial cues for each observed point: depth and motion across frames acquired over time.
Such cues can be obtained deploying two well-known techniques in computer vision: stereo matching and optical flow estimation. The first one aims at inferring the disparity (i.e. depth) by matching pixels across two rectified images acquired by synchronized cameras, the second at determining the 2D motion between corresponding pixels across two consecutive frames, thus requiring at least four images for full scene flow estimation.
For years, solutions to scene flow \cite{Behl2017ICCV} have been rather accurate, yet demanding in terms of computational costs and runtime. 
Meanwhile, deep learning established as state-of-the-art for stereo matching \cite{mayer2016large} and optical flow \cite{dosovitskiy2015flownet}. Thus, more recent approaches to scene flow leveraged this novel paradigm stacking together single-task architectures \cite{ilg2018occlusions,Ma_2019_CVPR}. However, this strategy is demanding as well and requires separate and specific training for each network and does not fully exploit the inherent dependency between the tasks, e.g. the flow of 3D objects depends on their distance, their motion and camera ego-motion \cite{Taniai_2017_CVPR}, as a single model could.
On the other hand, either synthetic or real datasets annotated with full scene flow labels are rare compared to those disposable for stereo and flow alone. This constraint limits the \textit{knowledge} available to a single network compared to the one exploitable by an ensemble of specialized ones.

\begin{figure}
    \centering
    
\begin{tabular}{c}
        \includegraphics[width=0.95\linewidth]{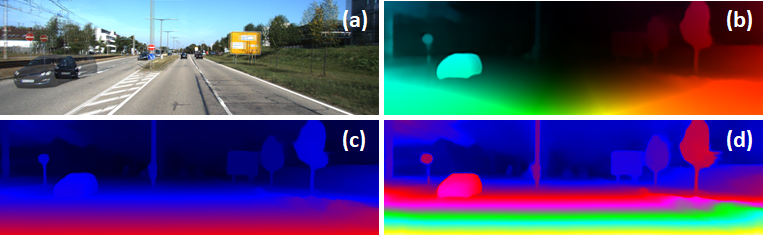} \\

    \end{tabular}
    \caption{\textbf{End-to-end scene flow with DWARF.} (a) Superimposed reference images at time $t_1$ and $t_2$, (b) estimated optical flow, (c) disparity and (d) disparity change.}
  \label{fig:abstract}
\end{figure}

To tackle previous issues, in this paper, we propose a novel lightweight architecture for scene flow estimation jointly inferring disparity, optical flow and disparity change (i.e., the depth component of 3D motion). We design a custom layer, namely 3D correlation layer, by extending the formulation used to tackle the two tasks singularly \cite{dosovitskiy2015flownet,mayer2016large}, in order to encode matching relationships across the four images. 
Moreover, to overcome the constraint on training data, we recover the missing knowledge leveraging standalone, state-of-the-art networks for stereo and flow to generate proxy annotations for our single scene flow architecture. Using this strategy on the KITTI dataset \cite{KITTI_2015}, we \textit{distill} about $20\times$ samples compared to the number of ground truth images available, enabling for more effective training and thus to more accurate estimations.

Our architecture for scene flow estimation through \underline{D}isparity, \underline{War}ping and \underline{F}low (dubbed \textbf{DWARF}) can be elegantly trained in an end-to-end manner from scratch and yields competitive results compared to state-of-the-art, although running about $10\times$ faster thanks to efficient design strategies. Figure \ref{fig:abstract} shows a qualitative example of dense scene flow estimation achieved by our network, enabling 10 FPS on NVIDIA Titan 1080Ti and about 1 FPS on Jetson TX2 embedded system. 

\section{Related Work}

We review the literature concerning deep learning for optical flow and stereo, as well as scene flow estimation.

\textbf{Optical flow.}
Starting from the seminal work by \citet{horn1981determining}, many others researchers mainly tackled optical flow deploying variational \cite{sun2014quantitative,epic_flow} and learning based \cite{learning_optical_flow,wulff2015efficient} approaches.
Nonetheless, starting from FlowNet \cite{dosovitskiy2015flownet} most recent works rely on deep learning. Specifically, it introduces the design of a 2D correlation layer encoding similarity between pixels, rapidly becoming a standard component in end-to-end networks for flow and stereo. The results obtained by FlowNet have been improved stacking more networks \cite{ilg2017flownet}, significantly increasing the number of parameters of the overall model. SpyNet \cite{Ranjan_2017_CVPR} addresses the complexity issue through coarse-to-fine optical flow estimation. 
PWCNet \cite{pwcnet} and LiteFlowNet \cite{hui18liteflownet} further improved this strategy using a correlation layer at each stage of the pyramid.
Finally, self-supervised optical flow has been studied by leveraging view synthesis \cite{meister2018unflow} or by distilling labels in a teacher-student scheme \cite{DDFlow}. 

\textbf{Stereo matching.}
Inferring depth from stereo pairs is a long-standing problem in computer vision, and well-known geometric constraints can be exploited to estimate disparity and then to obtain depth by triangulation.
Although traditional methods such as SGM \cite{hirschmuller2005accurate} are a popular choice, deep learning gave a notable boost in accuracy and it represents the state-of-the-art. \citet{zbontar2016stereo} replaced conventional matching costs computation with a siamese CNN network. \citet{luo2016efficient} cast the correspondence problem as a multi-class classification task. \citet{mayer2016large} introduced \textit{DispNetC}, an end-to-end trainable network leveraging on the same correlation layer proposed for flow \cite{dosovitskiy2015flownet}, but applied to 1D domain. \citet{Kendall_2017_ICCV} proposed to stack a cost volume and to process it with 3D convolutions. Following these latter two design strategies, many works have been proposed leveraging correlation scores \cite{Liang_2018_CVPR,yang2018segstereo,song2018stereo}, 3D convolutions \cite{Chang_2018_CVPR,zhang2019ga} or both \cite{guo2019group} to further improve the final accuracy. \citet{Poggi_CVPR_2019} improved end-to-end stereo with LiDAR guidance.
As for optical flow, coarse-to-fine strategies and warping \cite{Tonioni_2019_CVPR,yin2019hierarchical,dovesi2019real} yielded compact, yet accurate, architectures suited even for embedded devices.

\begin{figure*}
\centering
\includegraphics[width=0.95\textwidth]{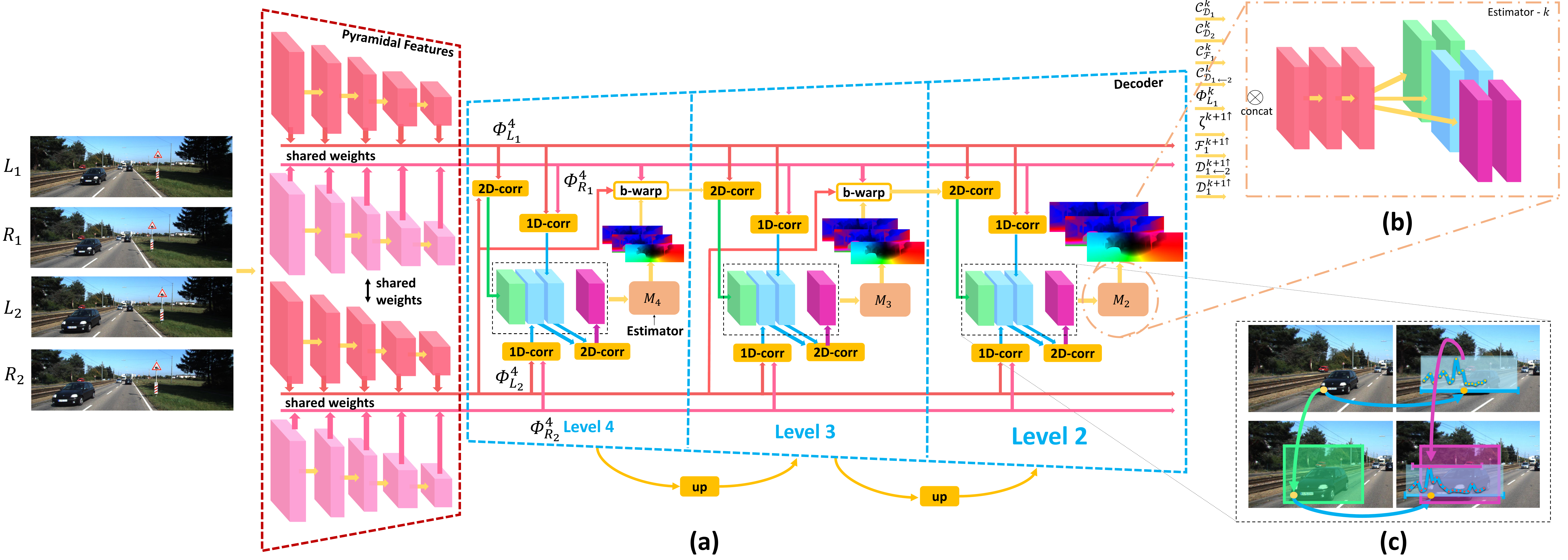}
\caption{\textbf{DWARF architecture.} The full architecture (a) has shared encoders (pink) to extract pyramids of features. At each resolution $k$, correlation scores respectively in green, light blue, light blue and purple, are stacked and forwarded to the estimator to generate $\mathcal{F}^k_1$, $\mathcal{D}^k_1$ and $\mathcal{D}^k_{1\leftarrow2}$. Such outputs are used to warp features at level $k-1$ until the final resolution is reached. Each estimator (b) is made of a common backbone followed by three task-specific heads. Correlation layers encode matching between pixels across the four images (c).
}
\label{fig:architecture}
\end{figure*}

\textbf{Scene flow.} \citet{sceneflow} represents the very first attempt to estimate scene flow from multi-view frame sequences. Most times this task has been cast as a variational problem \cite{Bro11,Pons-ijcv,Basha-ijcv,Zhang-CVPR01,huguet2007variational,valgaerts2010joint}. 
Some works applied 3D regularisation techniques based on a model of dense depth and 3D motion \cite{vogel20113d} and a piecewise rigid scene model (better known as PRSM) for stereo \cite{vogel2013piecewise} and multi-frame \cite{vogel20153d}. \cite{Behl2017ICCV} involve object and instance recognition performed by a CNN into scene flow estimation within a CRF-based model,
\citet{KITTI_2015} tackled scene flow estimation by segmenting the scene into multiple rigidly-moving objects.
\citet{Taniai_2017_CVPR} deployed a pipeline for multi-frame scene flow estimation, visual odometry and motion segmentation running in a couple of seconds.
Concerning deep learning methods, \citet{ilg2018occlusions} trained specialised networks for stereo and optical flow and combined them within a refinement module to estimate disparity change. Similarly, \citet{Ma_2019_CVPR} rely on pre-existing networks for stereo \cite{Chang_2018_CVPR}, optical flow  \cite{pwcnet} and segmentation \cite{He_2017_ICCV} and then infer scene flow through a Gaussian-Newton solver.
Despite much faster compared to the prior state-of-the-art, they require multi-stage training protocols for every single task and power-hungry GPUs to barely run at 1 FPS. 

Recently, \citet{saxena2019pwoc} proposed a
fast and lightweight model taking into account occlusions. Although similar in design, we will show that DWARF outperforms it by a good margin. Concurrent to our work, \citet{Jiang_2019_ICCV} propose a similar architecture to jointly learn for scene flow and semantics.

\section{Proposed Architecture}

In this section, we introduce the DWARF architecture built upon established principles from optical flow and stereo matching to obtain, in synergy, an end-to-end framework for full scene flow estimation. As already proved in different fields, coarse-to-fine designs enable for compact, yet accurate models.  

Given a couple of stereo image pairs $L_{1}, R_{1}, L_{2}$ and $R_{2}$ referencing, respectively, the left and right images at time $t_{1}$ and $t_{2}$ we aim at estimating disparity $\mathcal{D}_1$ between $L_1,R_1$ to obtain its 3D position at time $t_1$, optical flow $\mathcal{F}_1$ between $L_1,L_2$ to get 2D motion vectors connecting pixels in $L_1$ to those in $L_2$ and disparity change $\mathcal{D}_{1\leftarrow2}$, i.e. disparity $\mathcal{D}_{2}$ between $L_2,R_2$ mapped on corresponding pixels in image $L_1$ that allows to get $z$ component of 3D motion vectors.
To achieve this, our model performs a first extraction phase in order to retrieve a pyramid of features from each image, then in a coarse-to-fine manner it computes point-wise correlations across the four features representations and estimates the aforementioned disparity and motion vectors, going up to the last level of the pyramid to obtain the final output. Figure \ref{fig:architecture} sketches the structure of DWARF configured to process, for the sake of space, a pyramid down to $\frac{1}{32}$ of the original resolution. In the next section, we will describe in detail each module depicted in the figure.

\subsection{Features Extraction}  

To extract meaningful representations from each input image, we design a compact encoder to obtain a pyramid of features ready to be processed in a coarse-to-fine manner. Purposely, DWARF has four encoders, one for each input image, with shared weights. Each one is built of a block of three $3\times3$ convolutional layers for each level in the pyramids of features, respectively with stride 2, 1 and 1. For the sake of space, Figure \ref{fig:architecture} (a) shows an example of 5 levels encoder. Actually DWARF deploys a 6 levels encoder down to $\frac{1}{64}$ resolution features ($k$=$6$), counting 18 convolutional layers, each followed by Leaky ReLU activations with $\alpha=0.1$. By progressively decimating the spatial dimensions, we increase the amount of extracted features, respectively to 16, 32, 64, 96, 128 and 196. 
It generates features $\phi^k_{L_{1}}, \phi^k_{R_{1}}, \phi^k_{L_{2}}$ and $\phi^k_{R_{2}}$ with $k \in [1,6]$, respectively for frames $L_{1}, R_{1}, L_{2}$ and $R_{2}$, deployed by the following module to extract matching relationships between pixels.

\subsection{Warping}
\label{sec:warping}

The main advantage introduced by a coarse-to-fine strategy consists of computing small disparity and flow vectors at each resolution and sum them while going up the pyramid. This strategy allows keeping a small range where to calculate correlation scores, as we will discuss in detail in the next section. Otherwise, a large search space would dramatically increase the complexity of the entire network.

Given features $\phi^k_{L_{1}}, \phi^k_{R_{1}}, \phi^k_{L_{2}}$ and $\phi^k_{R_{2}}$ extracted by the encoder at the $k^{th}$ level, we have to bring all features closer to $\phi^k_{L_{1}}$ coordinates. To do so, estimates at previous pyramid level $(k+1)$ are upsampled, e.g. $\mathcal{D}^{k+1}_1$ to $\mathcal{D}^{{k+1}^\uparrow}_{1}$, and properly scaled to match stereo/flow at the next resolution $k$. Then, features are warped by means of backward warping, in particular $\phi^k_{L_{2}}$ according to optical flow $\mathcal{F}^{{k+1}^\uparrow}_{1}$ and $\phi^k_{R_{1}}$ according to $\mathcal{D}^{{k+1}^\uparrow}_{1}$.
Finally, the motion that allows to warp $\phi^k_{R_{2}}$ towards $\phi^k_{L_{1}}$ is given by the sum of $\mathcal{F}^{{k+1}^\uparrow}_{L_{1}}$ and $\mathcal{D}^{{k+1}^\uparrow}_{1\leftarrow2}$. This because the former encodes the mapping between present and future correspondences, while the latter the horizontal displacement occurring between $L_2$ and $R_2$, but on $L_1$ coordinate, thus the same as $\mathcal{F}^{{k+1}^\uparrow}_1$.

We will see how this translates into computing, at each resolution $k$, a refined scene flow field to ameliorate a prior, coarse estimation inferred at resolution $k+1$.
At the lowest resolution in the pyramid, features are not warped since scene flow priors are not available. 

\subsection{Cost Volumes and 3D Correlation Layer}
\label{sec:3dcorr}

Since DWARF jointly reasons about stereo and optical flow, correlation layers fit very well in its design.
At first we compute correlation scores encoding standalone tasks, i.e. estimation of disparity $\mathcal{D}_{1}$ between $L_1,R_1$, $\mathcal{D}_{2}$ between $L_2,R_2$ and flow $\mathcal{F}_{1}$ between $L_1,L_2$, obtaining $\mathcal{C}^k_{\mathcal{D}_{1}}(\mathcal{\phi}^k_{L_1},\mathcal{\phi}^k_{R_1})$, $\mathcal{C}^k_{\mathcal{D}_{2}}(\mathcal{\phi}^k_{L_2},\mathcal{\phi}^k_{R_2})$, $\mathcal{C}^k_{F_{1}}(\mathcal{\phi}^k_{L_1},\mathcal{\phi}^k_{L_2})$ by means of two 1D and one 2D correlation layers depicted in light blue and green in Figure \ref{fig:architecture} (a). By defining the correlation between per-pixel features as $\langle \cdot \rangle$ and concatenation as $\otimes$, we obtain 1D and 2D correlations as

 \begin{equation}
 \scriptsize
 \begin{split}
    &\mathcal{C}^k_{\mathcal{D}_{t}}(y,x) = \bigotimes_{j \in [-r_x,r_x]} \langle \phi^k_{L_{t}}(y,x), \phi^k_{R_{t}}(y,x+j)_w \rangle \\
    &\mathcal{C}^k_{\mathcal{F}_{1}}(y,x) = \bigotimes_{\substack{i \in [-r_y,r_y],\\j \in [-r_x,r_x]}} \langle \phi^k_{L_{1}}(y,x), \phi^k_{L_{2}}(y+i,x+j)_w \rangle
 \end{split}
 \end{equation}
with $(y,x)$ pixel coordinates, $r_y,r_x$ radius on $y$ and $x$ directions, $t \in [1,2]$. Subscript $w$ means warping via upsampled priors as described in Section \ref{sec:warping}.
Although such features embody relationships about standalone tasks, they lack at encoding matching between the 3D motion of the scene. To overcome this limitation, we introduce a novel custom layer.

Figure \ref{fig:architecture} (c) depicts how correlation layers act in DWARF. While 2D correlation layer (green) encodes similarities between pixels aimed at estimating optical flow, 1D correlations (light blue) compute scores between left and right images independently from time. Each produces a correlation curve, superimposed on $R_1$ and $R_2$ in the figure. If a pixel does not change its disparity through time, the peaks in the two curves would ideally match. Otherwise, they will appear shifted by the magnitude of the disparity change. 
The rest of the curve will shrink/enlarge, with major differences in portions dealing with regions moving of different motions (e.g., background vs foreground objects). 
This pattern, if properly learned, acts as a bridge between depth and 2D flow, enabling to infer the full 3D motion.
Unfortunately, this behaviour is not explicitly modelled by the layers mentioned above. Hence, we adopt a novel component, namely a 3D correlation layer, whose search volume is depicted in purple in Figure \ref{fig:architecture} (c). Since correlation curves are already available from 1D correlation layers, this translates into computing \emph{correlations over correlations} volumes as

\begin{equation}
\scriptsize
\mathcal{C}^k_{\mathcal{D}_{1\leftarrow2}} = \bigotimes_{\substack{i \in [-r_y,r_y],\\j \in [-r_x,r_x],\\h \in [-r_z,r_z]}} \langle \mathcal{C}^k_{\mathcal{D}_{1}}(y,x,d), \mathcal{C}^k_{\mathcal{D}_{2}}(y+i,x+j,d+h) \rangle
\end{equation}

with $(y,x,d)$ pixel coordinates in the correlation volumes and $r_z$ the search radius for displacement between 1D correlation curves.
Specifically, the full search space of such operation is 3D, being it over pixel coordinates plus displacement between correlation curves. We refer the reader to the \textbf{supplementary material} for some examples supporting this rationale, while ablation studies reported among our experiments prove the effectiveness of such a new layer, allowing DWARF to outperform similar architectures \cite{saxena2019pwoc} on the KITTI online benchmark.

\subsection{Scene Flow Estimation}

After the extraction of meaningful correlation features, we stack them into a features volume forwarded to a compact decoder network in charge of estimating the three components of the scene flow. As shown in Figure \ref{fig:architecture} (b), at each level the volume contains reference image features $\phi_{L_0}^k$, correlation scores, upsampled scene flow priors and latest features $\zeta$ extracted before estimation at level $k+1$.
This input is forwarded to level $k$ decoder. First, three convolutional layers with respectively 128, 128 and 96 channels rearrange the volume. Then, three independent \textit{heads} are in charge of predicting $\mathcal{D}^k_{1}$, $\mathcal{F}^k_{1}$ and $\mathcal{D}^k_{1\leftarrow2}$. Following this design, the network is forced to create a first holistic representation of the volume, then specialized by each sub-module.
Each head has two task-specific $3\times3$ convolutional layers with 64 and 32 channels, producing $\zeta$ features from which a final $3\times3$ layer extracts the single component of scene flow at level $k$, e.g. $\mathcal{D}^k_{1}$. Such estimates, together with features $\zeta$, are upsampled through a transposed convolution layer with stride 2, to provide coarse scene flow priors for warping at level $k-1$. Leaky ReLU units with $\alpha=0.1$ follow all layers.
Each estimator is optionally designed with dense connections \cite{huang2017densely} to boost accuracy. This design choice adds about 10 million parameters to DWARF. 

\subsection{Residual Refinement}\label{sec:refinement}

Although the explicit reasoning about features matching across the four images is an effective way to guide the network towards scene flow estimation, it has limitations for pixels having missing correspondences.
This fact occurs when, in one or multiple frames, they are occluded or no longer part of the observed scene. For instance, portions of the sensed scene located near image borders at time $t_1$ are no longer framed at $t_2$ when the camera is moving.
To soften this problem, three residual networks are deployed to refine each single component of the full scene flow estimates, taking as input $\zeta$ features from the top-level estimator and processing it with six $3\times3$ convolutional layers extracting respectively 128, 128, 128, 96, 64, 32 features, with a dilation factor of 1, 2, 4, 8, 16, and 1 respectively to increase the receptive field introducing moderate overhead. A Leaky ReLU with $\alpha=0.1$ follows each layer. Then, a further $3\times3$ convolutional layer (without activation units) extracts residual scene flow, summed to previous final estimations in order to refine them.

\begin{figure}
    \centering
    \includegraphics[width=0.95\linewidth]{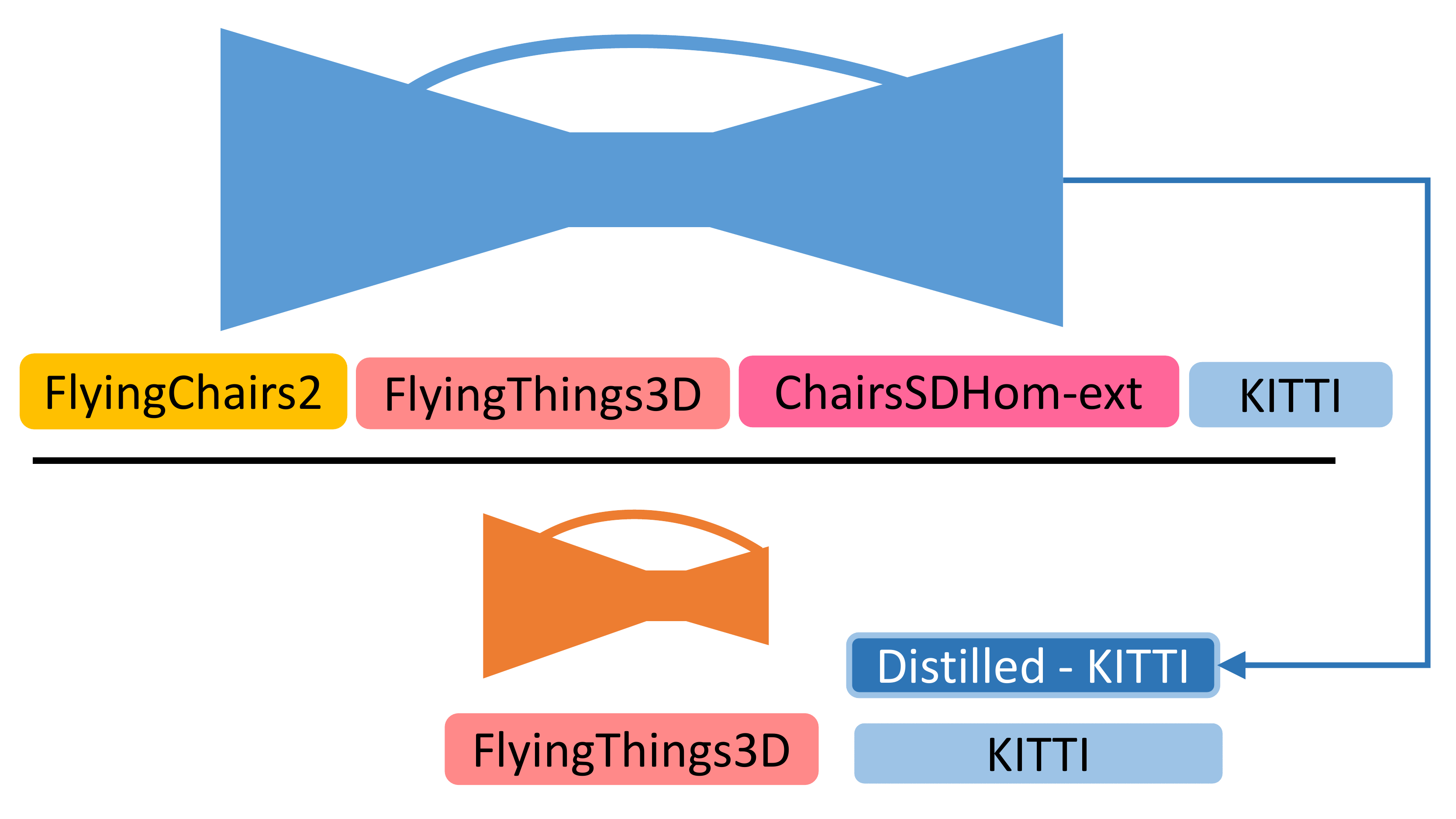}
    \caption{\textbf{Knowledge distillation \cite{hinton2015distilling} scheme.} From an ensemble of deep networks \cite{ilg2018occlusions} (blue) trained on a variety of datasets we transfer knowledge to our compact model (orange).}
    \label{fig:knowledge_distillation}
\end{figure}

\subsection{Knowledge Distillation from Expert Models}

As previously pointed out, although end-to-end training is elegant and easier to schedule, it prevents using task-specific datasets since ground truth labels are required for full scene flow. However, a proper training schedule across several datasets is needed to achieve the best accuracy on single tasks \cite{ilg2017flownet,pwcnet}. 
To overcome this limitation, we leverage on knowledge distillation \cite{hinton2015distilling} employing expert models trained for the single tasks and used to teach to a student network, DWARF in this case. 

Specifically, we choose the ensemble of networks proposed by \citet{ilg2018occlusions} to guide our simpler model, thanks to the availability of the source code and its excellent performance. Firstly a FlowNet-CSS and DispNet-CSS are in charge of optical flow and disparity estimation, then a third FlowNet-S architecture processes disparity $\mathcal{D}_2$ back warped according to computed optical flow and refines it to obtain $\mathcal{D}_{1\leftarrow2}$. The three networks are trained in a multi-stage manner, starting from DispNet-CSS and FlowNet-CSS, and ending with the training of the final FlowNet-S. This allows for multi-dataset training, especially in the case of optical flow for which several sequential rounds of training on FlyingChairs2 and ChairsSDHom-ext \cite{ilg2017flownet} are performed to achieve the best accuracy.
By teaching DWARF with the expert models, we are able to both i) bring the knowledge learned by the expert model on task-specific datasets (e.g.,  FlyingChairs2 and ChairsSDHom-ext) to our model and ii) distill an extended training set, counting a larger amount and more variegated samples.
We will show in our experiments how the knowledge distillation scheme results more effective than training on the few ground truth images available from real datasets.

\subsection{Training Loss}\label{training_loss}

Given the set of learnable parameters of the network $\Theta$, $\mathcal{D}^k_{1}(\Theta)$, $\mathcal{D}^k_{1\leftarrow2}(\Theta)$ and $\mathcal{F}^k_{1}(\Theta)$ respectively the estimated disparity, disparity change and optical flow, $\mathcal{D}^k_{1}(GT)$, $\mathcal{D}^k_{1\leftarrow2}(GT)$ and $\mathcal{F}^k_{1}(GT)$ the ground truth maps for specific scene flow components brought to each pyramid level $k$, we adopt the L1 norm to optimise DWARF:
 
 \begin{equation} \label{loss}
 \scriptsize
 \begin{split}
    &\mathcal{L}(\Theta)=\gamma \|\Theta\|^2_{1} + \epsilon_{1}\sum_{k=l_{1}}^L \alpha_{k} \|\mathcal{D}_{1}(\Theta)-\mathcal{D}^k_{1}(GT)\|_1 \\
    +&\epsilon_{2} \sum_{k=l_{1}}^L \alpha_{k} \|\mathcal{D}^k_{1\leftarrow2}(\Theta)-\mathcal{D}^k_{1\leftarrow2}(GT)\|_1 +
    \epsilon_{3} \sum_{k=l_{1}}^L \alpha_{k} \|\mathcal{F}^k_{1}(\Theta)-\mathcal{F}^k_{1}(GT)\|_1 
 \end{split}
 \end{equation}
We deploy a 6 levels pyramidal structure, extracting features up to level 6, halving the spatial resolution down to $\frac{1}{64}$. We set $l_0=2$, thus estimating scene flow up to quarter resolution and then bilinearly upsampling to the original input resolution. This strategy allows us to keep low memory requirements and fast inference time. The search spaces are set to $9$, $9\times9$ and $9\times9\times1$ respectively for 1D, 2D  and 3D correlations. A search range of 1 on disparity change keeps low the overall complexity of the network, yet significantly improving the accuracy on all metrics.

\section{Experimental Results}

We report extensive experiments aimed at assessing the accuracy and performance of DWARF. First, we describe in detail the training schedules.
Then, we conduct an ablation study to measure the contribution of each component and compare DWARF to state-of-the-art deep learning approaches. Finally, we focus on DWARF run-time performance, extensively studying its behaviour on a variety of hardware platform, including a popular embedded device equipped with a low-power GPU.

\subsection{Training Datasets and Protocol}

It is a common practice to initialize end-to-end networks on large synthetic datasets before fine-tuning on real data \cite{dosovitskiy2015flownet,mayer2016large,ilg2017flownet}.
Despite the large availability of synthetic datasets for flow and stereo \cite{sintel,dosovitskiy2015flownet,mayer2016large}, only the one proposed in \cite{mayer2016large} provides ground truth for full scene flow estimation. In this field, KITTI 2015 \cite{KITTI_2015} represents the unique example of a realistic benchmark for scene flow. Therefore, we scheduled training on these two datasets and, optionally, we leverage knowledge distillation \cite{hinton2015distilling} from an expert network \cite{ilg2018occlusions} to augment the variety of realistic samples and consequently to better train DWARF.

\textbf{Flying Things 3D.}
We set $\alpha_{6}=$ 0.32, $\alpha_{5}=$ 0.08, $\alpha_{4}=$ 0.02, $\alpha_{3}=$ 0.01 and $\alpha_{2}=$ 0.005, $\gamma=$ 0.0004 and cross-task weights to $\epsilon_{1}=$ 1, $\epsilon_{2}=$ 1 and $\epsilon_{3}=$ 0.5.
Ground truth values are down-scaled to match the resolution of the level and scaled by a factor of 20, as done by \citet{dosovitskiy2015flownet,pwcnet}. 
The network has been trained for $1.2$M steps with a batch size of 4 randomly selecting crops with size 768$\times$384, using Adam optimiser \cite{kingma2014adam}, with $\beta_{1}=0.9$, $\beta_{2}=0.999$  and initial learning rate of $10^{-4}$, which has been halved after $400$K, $600$K, $800$K and $1$M steps.

\textbf{KITTI 2015.}  \label{KITTI_ablation}
We fine-tuned the network using the 200 training images from the KITTI Scene Flow \cite{Menze2015ISA} dataset with a batch size of 4 for 50K steps. Again, Adam optimizer \cite{kingma2014adam} has been adopted with the same parameters as before. The initial learning rate is set to $3\times10^{-5}$, halved after 25K, 35K and 45K steps. We minimise loss only at level $k=2$. Specifically, we upsample through bilinear interpolation the predictions at the quarter resolution and apply the fine-tuning loss described in \ref{training_loss} at full resolution. Predictions at lower levels have not been optimized explicitly.  
We set $\epsilon_{1}=$ 1, $\epsilon_{2}=$ 1 and $\epsilon_{3}=$ 0.5, all the $\alpha_{k}$ set to 0 with the exception of $\alpha_{2}$ set to 0.001 while $\gamma$ is left untouched. Images are firstly padded to $1280\times384$ pixels, then random crops of size $896\times320$ are extracted at each iteration. 

\textbf{Distilled-KITTI.} Finally, we perform knowledge distillation to produce an extended set of images for fine-tuning DWARF. Specifically, we use the 4000 total images available from the multiview extension of the KITTI 2015 training set and we produce proxy annotations leveraging FlowNet-CSS, DispNet-CSS and FlowNet-S. We use the trained models made available by the authors, trained on multiple task-specific datasets and fine-tuned on the aforementioned KITTI 2015 split (i.e., 200 images).
We point out that, excluding the task-specific synthetic datasets, the expert models are trained with the same real ground truth (i.e., no additional annotations) and are used only to distill more \emph{proxy} labels. Moreover, despite the extremely accurate estimates produced by the expert models on the KITTI training split (below 2\% error rate on full scene flow), the labels sourced through distillation are yet \emph{noisy}.

\textbf{Data augmentation.}
We perform data augmentation by applying random gamma correction in [0.8,1.2], additive brightness in [0.5,2.0], and colour shifts in [0.8,1.2] for each channel separately. To increase robustness against brightness changes, we applied augmentation independently to every single image.
Instead, random zooming, with probability 0.5 of re-scaling the image by a random factor in [1,1.8], has been applied in the same way to $L_{1}, R_{1}, L_{2}$ and $R_{2}$ and the relative ground truths.

\subsection{Ablation Studies}\label{ablation}

In this section, we study the effectiveness of each architectural choice.
Tables \ref{tab:syntethic} and \ref{tab:kitti} report experimental results on FlyingThings3D \cite{mayer2016large} test set and KITTI 2015 training set \cite{Menze2015ISA} by i) increasing the complexity of the network and ii) introducing the knowledge distillation process. 

\begin{table} 
\centering
\resizebox{.95\columnwidth}{!}{
\begin{tabular}{ccc|c|ccc}
\toprule
\multicolumn{3}{c|}{Configuration} & Params & \multicolumn{1}{c}{Flow} & \multicolumn{1}{c}{Disparity} & \multicolumn{1}{c}{Change} \\
\hline
Dense & 3Dcorr & Refine & M  & EPE & EPE & EPE  \\
\hline
&  & & 5.06 &  7.435 & 1.959 & 2.283 \\
\checkmark & & & 13.50 & 6.758 & 1.837 & 2.092 \\
\checkmark & \checkmark & &  15.87 & 6.738 & 1.827 & 2.149 \\
\checkmark & \checkmark & \checkmark & 19.62 & \textbf{6.440} & \textbf{1.784}  & \textbf{2.039} \\
\hline
\end{tabular}
}
\caption{\textbf{Ablation study on the FlyingThings3D test set.} For each variant of DWARF, we report End Point Error (EPE) for flow, disparity and change respectively.}
\label{tab:syntethic}
\end{table}

\begin{figure}
    
	\begin{tabular}{c}
    	\includegraphics[width=0.95\linewidth]{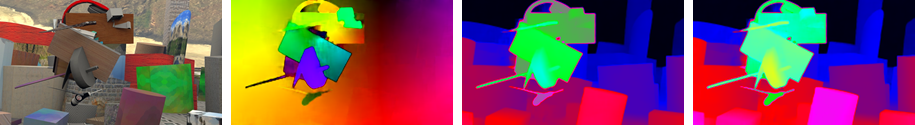} \\
    \end{tabular}
    \caption{\textbf{Qualitative results on FlyingThings 3D \cite{mayer2016large} test split}. From left to right, reference image at $t_1$, $\mathcal{F}_1$, $\mathcal{D}_1$ and $\mathcal{D}_{1\leftarrow 2}$.}
  \label{fig:flying}
  \end{figure}

\textbf{FlyingThings3D.} \label{Flying_ablation}
This dataset provides 
4248 frames for validation.
In Table \ref{tab:syntethic} we report average End-Point-Error (EPE) for the disparity, flow and change (respectively D1, F1 and D2) on 3822 images, obtained by filtering the validation set according to the guidelines.
We trained four DWARF variants, starting from the simple \textit{Standalone tasks} version (i.e., without Dense, 3D correlation and Refinement module) up to the full DWARF architecture.
We can notice how the addition of each module always yields better accuracy on most metrics. 
At first, adding dense connections improves over the baseline model at the cost of nearly triple the number of parameters. By introducing the 3D correlation module, we still improve the capability of the proposed solution to estimate the 3D motion of the scene, this time adding about 2M parameters to the previous 13.5. Finally, adding the Refinement module yields a consistent error reduction on all metrics. Figure \ref{fig:flying} shows qualitative results on FlyingThings3D validation set obtained by the full model. More qualitative examples are reported in \textbf{supplementary material.}

\textbf{KITTI 2015.}
For this experiment, we split the KITTI 2015 training set into 160 images for fine-tuning and reserve the last portion of 40 images for validation purposes only. We report F1, D1 and D2, respectively the percentages of pixels having absolute error larger than 3 and relative error larger than 5\% for the three tasks, considering only the non occluded regions (\textit{Noc}) and the whole image (\textit{All}).
In this ablation experiments, we aim at assessing the impact of the knowledge distillation protocol on the final accuracy of DWARF.
Purposely, we fine-tuned DWARF on two different datasets: i) the 160 images mentioned above from KITTI 2015 and ii) 3200 images from Distilled-KITTI, corresponding to the 160 sequences belonging to KITTI 2015 multiview extension, respectively reported in the first and the second rows in Table \ref{tab:kitti}. 
We can notice that proxy labels (Px) yield worse performance for optical flow and consequently for full scene flow while allow improving the accuracy for disparity estimation.
Combining the two approaches (i.e., replacing 160 images of the Distilled-KITTI dataset with the real available ground truth, third row) produces result close to using only distilled labels.
Finally, running a multi-stage fine-tuning made of 40k steps with proxy labels and further 10k with ground truth (ie, first learning from many yet noisy annotations and then focusing on few perfect labels) dramatically improves the results on optical flow and thus full scene flow, as shown in the fourth row of the table.

\begin{table} 
\centering
\resizebox{.95\columnwidth}{!}{
\begin{tabular}{cccc|c|c|c|c}
\toprule
\multicolumn{4}{c}{Configuration}& \multicolumn{4}{c}{} \\
\midrule
  Dense & 3Dcorr & Refine & Sup. & F1-All & D1-All & D2-All & SF-All \\
\midrule
    \checkmark & \checkmark & \checkmark & Gt & 18.53 & 4.58 & 9.32 & 20.85 \\
    \checkmark & \checkmark & \checkmark & Px & 20.71 & 3.94 & 9.14 & 23.07 \\
    \checkmark & \checkmark & \checkmark & Px + Gt & 20.47 & \textbf{3.91} & 9.43 & 23.01 \\
    \checkmark & \checkmark & \checkmark & Px $\rightarrow$ Gt & \textbf{16.75} & 4.22 & \textbf{8.26} & \textbf{19.00} \\
    \hline
    \end{tabular}
    }
    \caption{\textbf{Impact of knowledge distillation and its scheduling on 40 images from KITTI training set.} We report the percentage of pixels with error higher than 3 and 5\% respectively for flow, disparity, change and full scene flow.}
    \label{tab:kitti}
\end{table}

\begin{table}
    \centering
    \resizebox{.95\columnwidth}{!}{
    \begin{tabular}{ccc|ccc|c}
    \toprule
     \multicolumn{3}{c|}{Configuration} & \multicolumn{3}{c|}{Jetson TX2} & 1080 Ti \\
     Dense & 3DCorr & Refine & Max-Q & Max-P & Max-N & ($\approx$250W)\\
    \midrule
        & &  & 0.79s & 0.65s & 0.57s & 0.09s \\
        \checkmark & & & 1.26s  & 1.05s & 0.91s & 0.10s \\ 
        \checkmark & \checkmark &  & 1.47s & 1.22s & 1.06s & 0.11s \\
        \checkmark & \checkmark & \checkmark & 2.21s & 1.83s & 1.59s & 0.14s \\
    \hline
    \end{tabular}
    }
    \caption{\textbf{Runtime analysis} for different variants of DWARF on NVIDIA Jetson TX2 (using Max-Q, Max-P, Max-N configurations) and NVIDIA GTX 1080Ti. Time in seconds.}
    \label{tab:jetson}
\end{table}

\subsection{Run-time Analysis}
In Table \ref{tab:jetson}, we report the time required to process a couple of stereo images for all variants of DWARF using two different devices. For this purpose, we considered NVIDIA 1080Ti GPU and NVIDIA Jetson TX2, an embedded system equipped with a low-power GPU. The latter device can work with three increasing energy-consumption configurations: \textit{Max-Q} ($<$7.5W), \textit{Max-P} ($\approx$10W) and \textit{Max-N} ($<$15W). Even in its more complex configuration, our network can estimate in the \textit{Max-P} configuration the scene flow on KITTI (4 images padded at $1280\times384$) in less than 2s, draining about $\frac{1}{25}$ of the energy required by the 1080Ti.

\subsection{KITTI 2015 Online Benchmark}

Finally, Table \ref{tab:kitti_benchmark} reports results for DWARF and state-of-the-art solutions for scene flow, both traditional and based on deep learning. For the final submission, we included all the training data (4000 proxies, 200 ground truths). We followed a 50k (proxy) plus 5k (ground truth) schedule, halving the learning rate at 25K and 35K while reducing it by one quarter at 50K. Despite yielding lower accuracy compared to much more complex state-of-the-art architectures \cite{Ma_2019_CVPR,ilg2018occlusions}, our network allows us to achieve competitive results using $\sim100$M fewer parameters and running more than $10 \times$ faster.
Compared to approaches closer to ours \cite{saxena2019pwoc}, we can notice that our architecture is much more accurate on all metrics, with margins of about 2.58, 1.8 and 1.73\% respectively on F1-All, D1-All and D2-All, leading to a 2.91\% improved scene flow estimation, thanks to both 3D correlation layer and knowledge distillation introduced in this paper. Despite counting more than double parameters, DWARF runs almost at the same speed.
For a complete comparison with state-of-the-art algorithms, please refer to the KITTI 2015 online benchmark. At the time of writings, DWARF ranks 15$^{th}$.

\begin{table}
    \centering
    \resizebox{.95\columnwidth}{!}
    {
    \begin{tabular}{l|c|c|c|c|c|c}
    \toprule 
    Method &
    D1-All & D2-All & F1-All & SF-All &
    Params (M) &
    Runtime (s) \\
    \midrule
    \citet{Behl2017ICCV} & 4.46 & 5.95 & 6.22 & 8.08 & - & 600 \\
    (Vogel et al. 2015) & 4.27 & 6.79 & 6.68 & 8.97 & - & 300 \\
    \hline
    \citet{ilg2018occlusions} & 2.16 & 6.45 & 8.60 & 11.34 & 116.68 & 1.72 \\
    \citet{Ma_2019_CVPR} & 2.55 & 4.04 & 4.73 & 6.31 & 136.38 & 1.03\\
    \hline
    \citet{saxena2019pwoc} & 5.13 & 8.46 & 12.96 & 15.69 & 8.05 & 0.13 \\
    \textbf{DWARF (ours)} & 3.33 & 6.73 & 10.38 & 12.78 & 19.62 & 0.14 \\    
    \hline
    \end{tabular}
    }
    \caption{\textbf{Results on the KITTI 2015 online benchmark.} Results for \cite{ilg2018occlusions} from the original paper, since no longer available online. Runtime on NVIDIA 1080 Ti.}
    \label{tab:kitti_benchmark}
\end{table}	

\begin{figure}
    \centering
    \begin{tabular}{c}
    \includegraphics[width=0.95\linewidth]{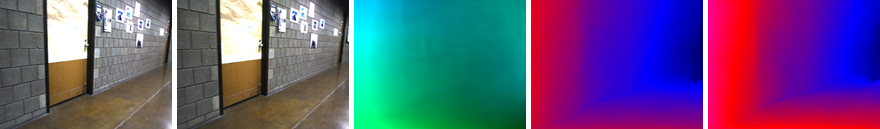} \\
    \end{tabular}
    \caption{\textbf{Qualitative results on the WeanHall dataset \cite{weanhall}.} From left to right, reference frames $L_1$ and $L_2$, $\mathcal{F}_1$, $\mathcal{D}_1$ and $\mathcal{D}_{1\leftarrow 2}$.}
    \label{fig:weanhall}
\end{figure}

\subsection{Additional Qualitative Results}

We also carried out additional experiments on the WeanHall dataset \cite{weanhall}, a collection of indoor stereo images. Since no ground truth is provided, we report qualitative results only. Figure \ref{fig:weanhall} depicts some examples extracted from this dataset processed by the same DWARF model used to submit results to the online KITTI benchmark, proving effective generalization to unseen indoor environments.
Finally, we refer the reader to \textbf{supplementary material} for additional qualitative results on both synthetic and real datasets, attached at the end of this manuscript. Moreover, a video is available at \UrlFont{\textbf{https://www.youtube.com/watch?v=qGWpi3z2M74}}.

\section{Conclusion}
In this paper, we proposed DWARF, a novel and lightweight architecture for accurate scene flow estimation. Instead of combining a stack of task-specialized networks as done by other approaches, our proposal is easily and elegantly trained in an end-to-end fashion to tackle all the tasks at once.
Exhaustive experimental results prove that DWARF is competitive with state-of-the-art approaches \cite{ilg2018occlusions,Ma_2019_CVPR}, counting 6$\times$ fewer parameters and running significantly faster.
Future work aims at self-adapting DWARF in an online manner \cite{Tonioni_2019_CVPR,Tonioni_2019_CVPRa}.

\textbf{Acknowledgements.} We gratefully acknowledge the
support of NVIDIA Corporation with the donation of the
Titan Xp GPU used for this research.

{\small
\bibliographystyle{aaai}
\bibliography{bibliography}
}

\setcounter{secnumdepth}{2}
\renewcommand\thefigure{\arabic{figure}}    
\setcounter{figure}{0}
\setcounter{section}{0}

\setlength{\droptitle}{-0.75in}
\title{\textbf{Learning End-To-End Scene Flow by Distilling Single Tasks Knowledge \\ -- Supplementary material}}
\author{}
\date{}

\maketitle

This document provides additional details concerned with AAAI 2020 paper ``Learning end-to-end scene flow by distilling single tasks knowledge". First, we discuss the fundamentals behind the design of our 3D correlation layer by showing a real example, then we report additional qualitative results on both synthetic and real datasets.

\begin{figure}
    \centering
    \begin{tabular}{cc}
        \includegraphics[width=0.45\textwidth]{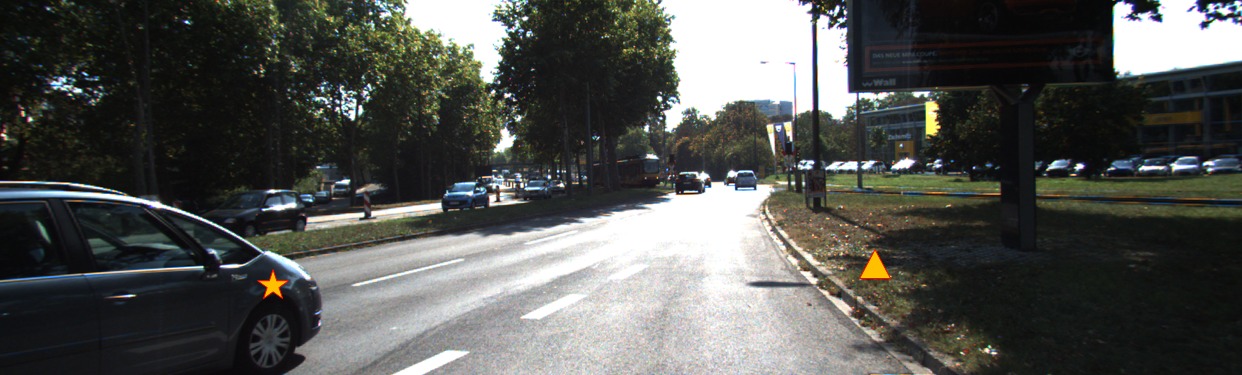} \\
        \includegraphics[width=0.45\textwidth]{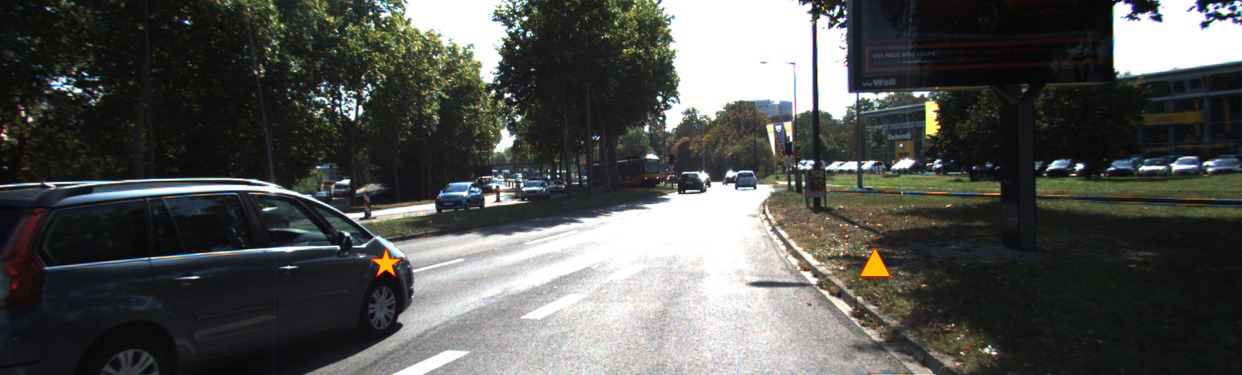} \\
    \end{tabular}
    \caption{Images 000091 from KITTI 2015, respectively at times $t_1$ and $t_2$. We highlight two pixels on both: one belonging to a static region (triangle) and one to a moving object (star).}
    \label{fig:star}
\end{figure}

\section{3D Correlation layer -- fundamentals} 

In this section, we show the rationale behind our 3D correlation layer with an example. Figure \ref{fig:star} depicts two left images from KITTI 2015 training set, $000091\_10$ on top and $000091\_11$ on bottom, acquired respectively at $t_1$ and $t_2$. Both are acquired by a static camera, thus the background is static with respect to it. We select two pixels in the first image, belonging to the background (triangle) or to a moving car (star), and their correspondences on the frame at $t_2$ according to ground truth optical flow. We can notice that the pixel on the car is getting away from the camera, i.e. the car is moving forward on the road, resulting in different disparities at $t_1$ and $t_2$. In particular, the latter will be minor than the former, being disparity proportional to inverse depth.

\begin{figure}
    \centering
    \begin{tabular}{cc}
        \includegraphics[width=0.18\textwidth]{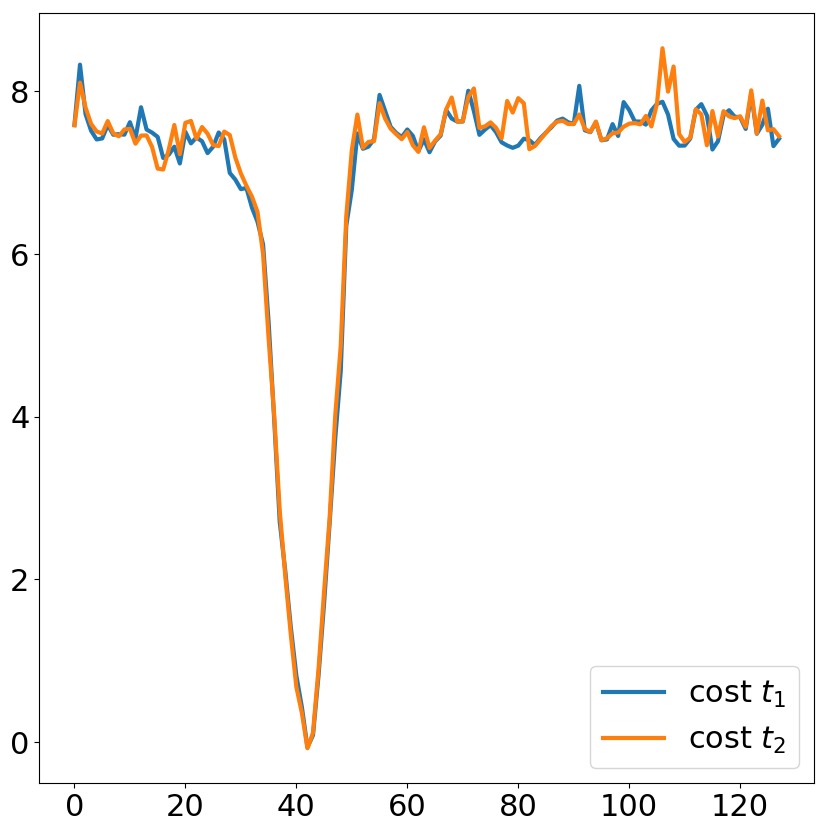} &
        \includegraphics[width=0.18\textwidth]{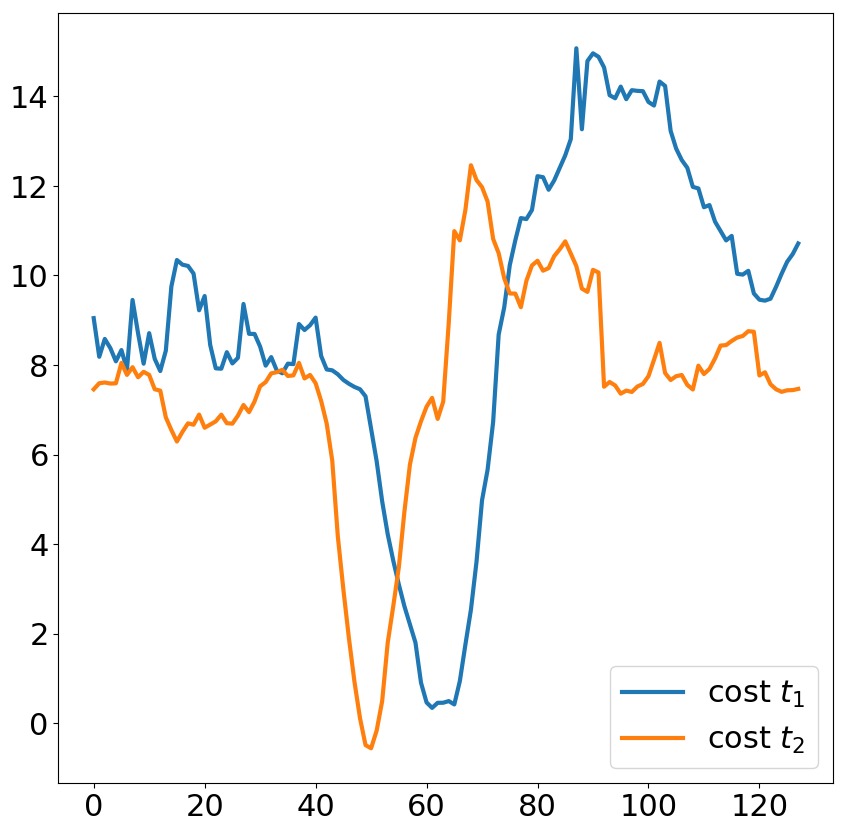} \\
    \end{tabular}
    \caption{Matching cost curves for static (left) and dynamic pixels (right), respectively at time $t_1$ (blue) and $t_2$ (orange). }
    \label{fig:match}
\end{figure}

\begin{figure}
    \centering
    \begin{tabular}{cc}
        \includegraphics[width=0.18\textwidth]{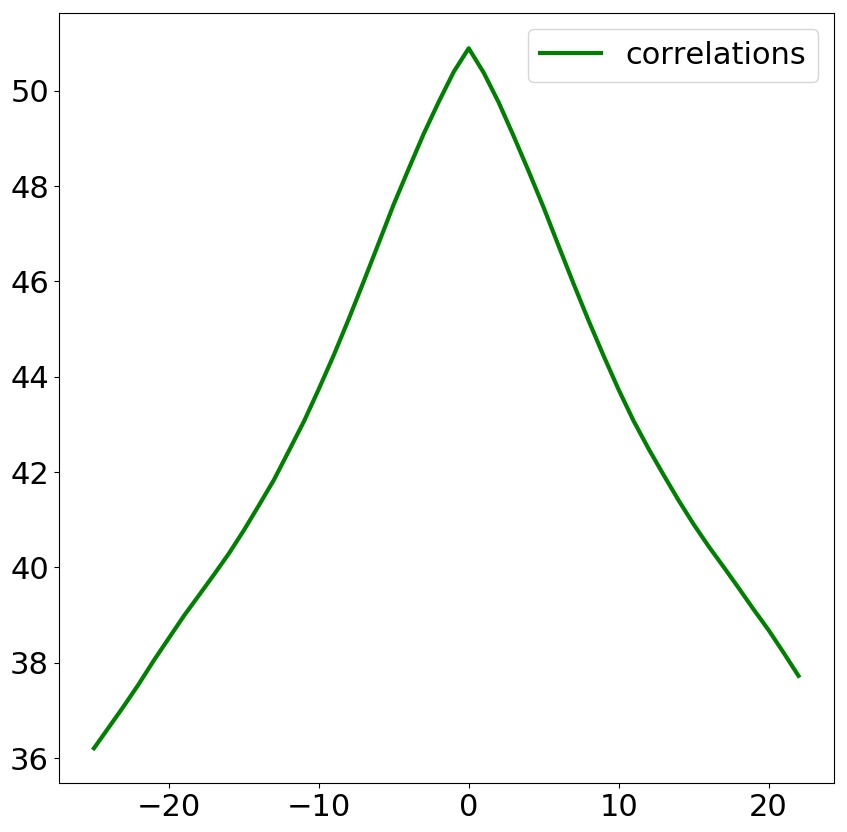} &
        \includegraphics[width=0.18\textwidth]{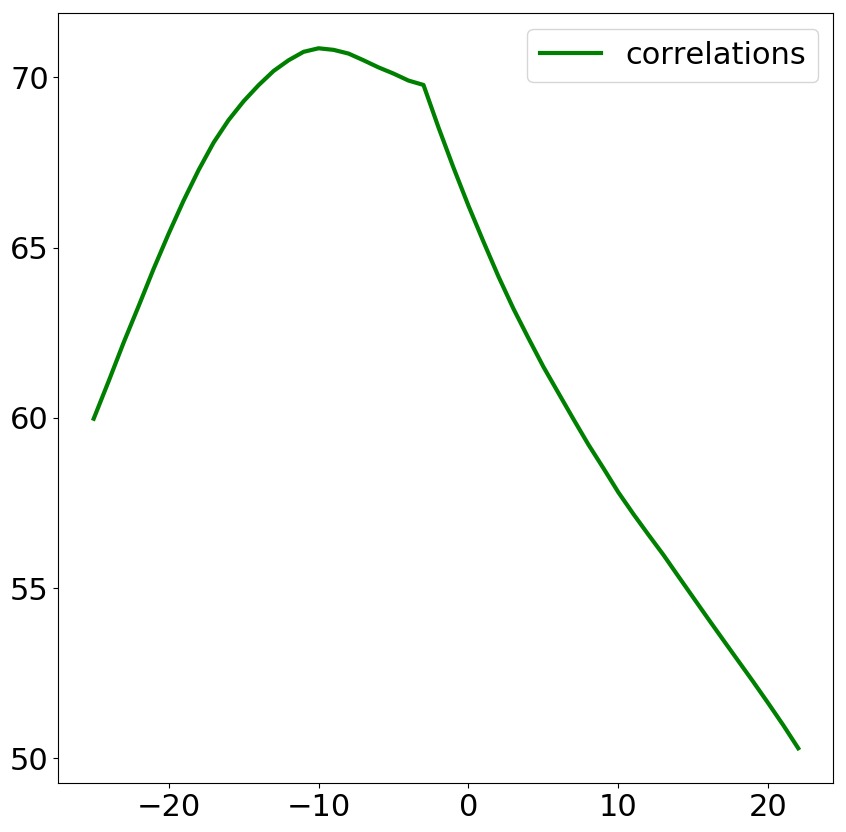} \\
    \end{tabular}
    \caption{Correlation curve for static (left) and dynamic (right) pixels between matching curves plotted in Figure \ref{fig:match}. }
    \label{fig:corrs}
\end{figure}

\begin{figure*}
    \centering
    \renewcommand{\tabcolsep}{1pt}
    \begin{tabular}{cccc}
    
        \includegraphics[width=0.22\textwidth]{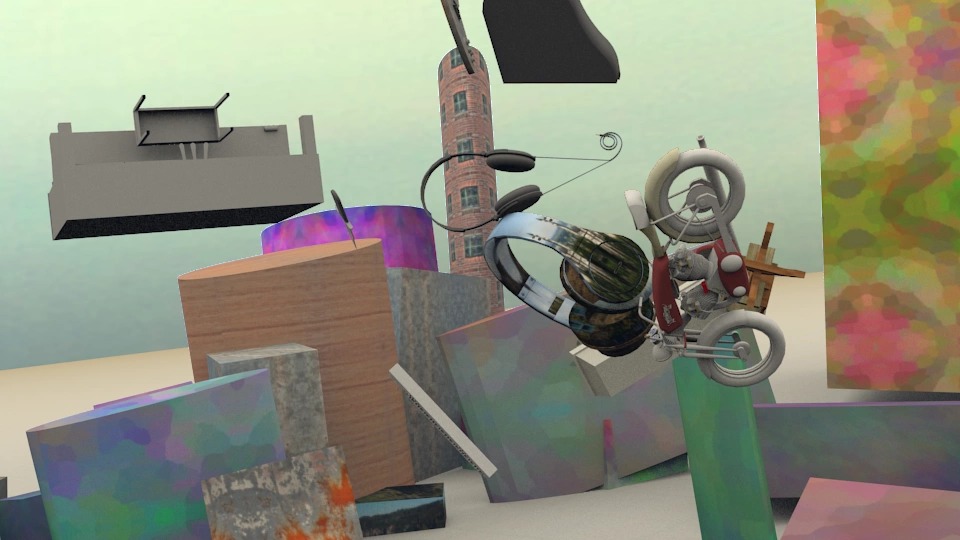} & \includegraphics[width=0.22\textwidth]{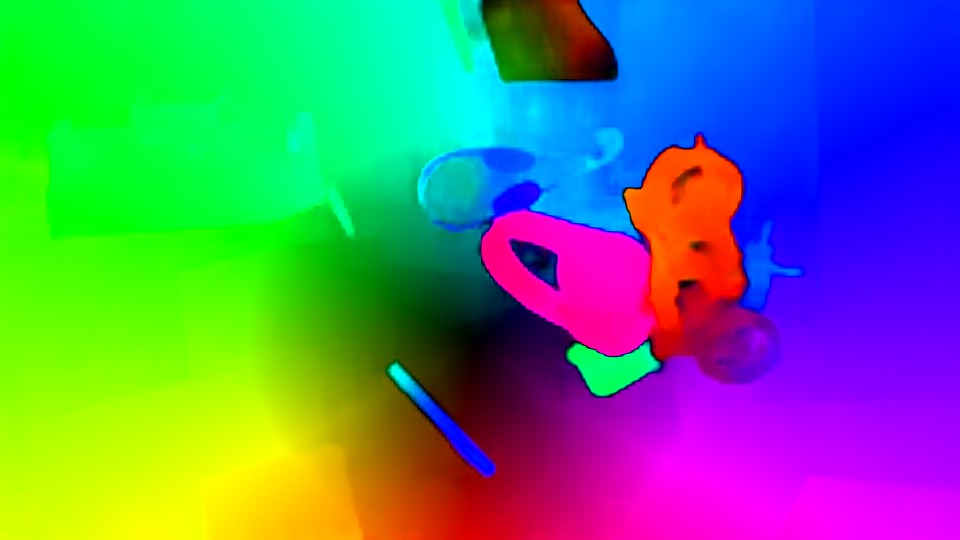} &
        \includegraphics[width=0.22\textwidth]{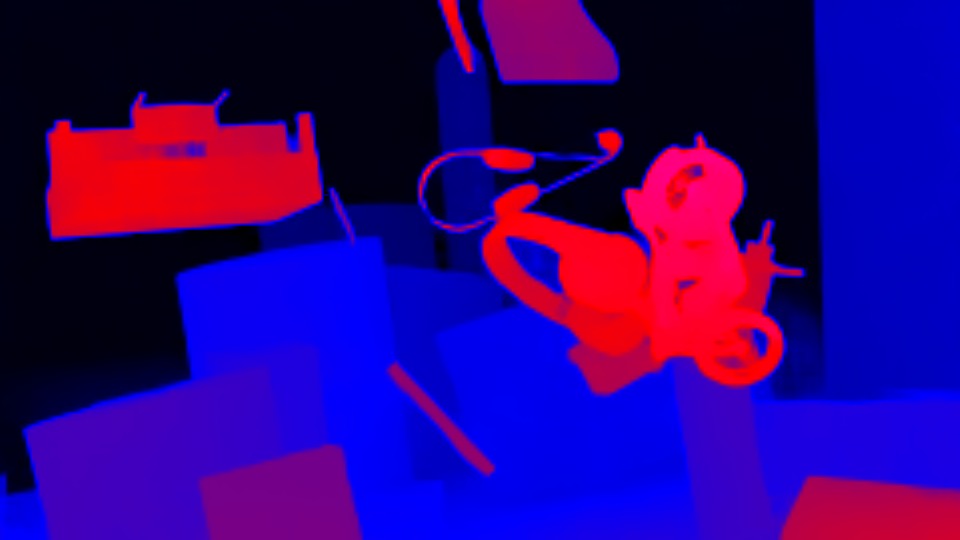} &
        \includegraphics[width=0.22\textwidth]{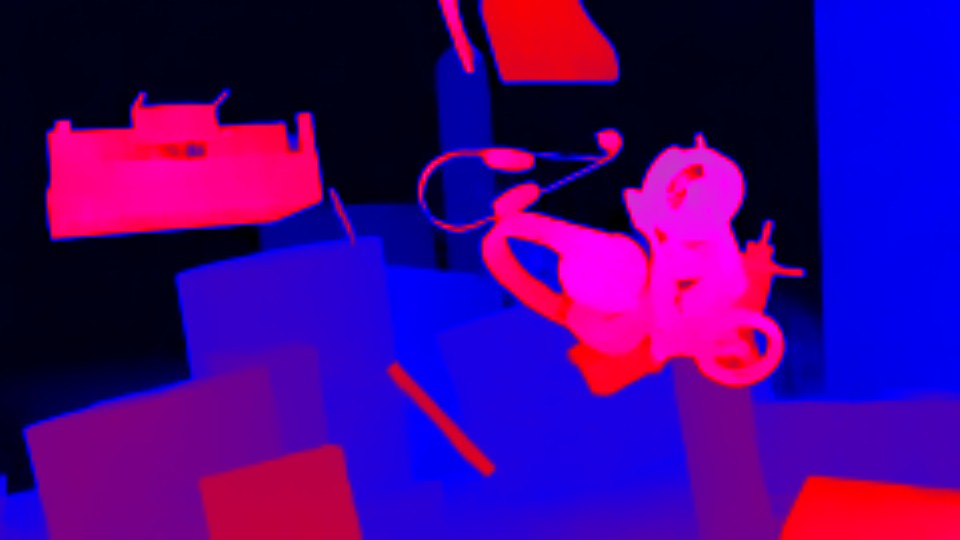} \\     
    
        \includegraphics[width=0.22\textwidth]{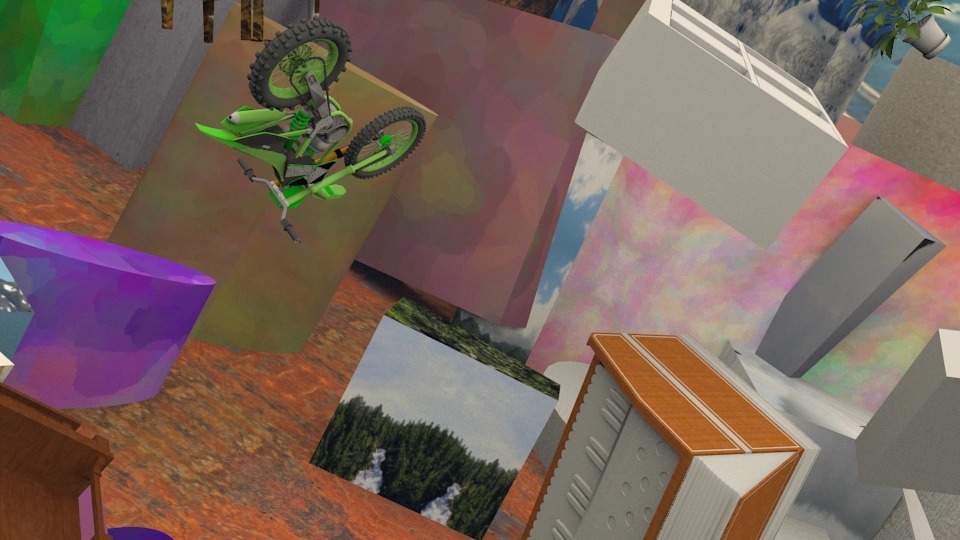} & \includegraphics[width=0.22\textwidth]{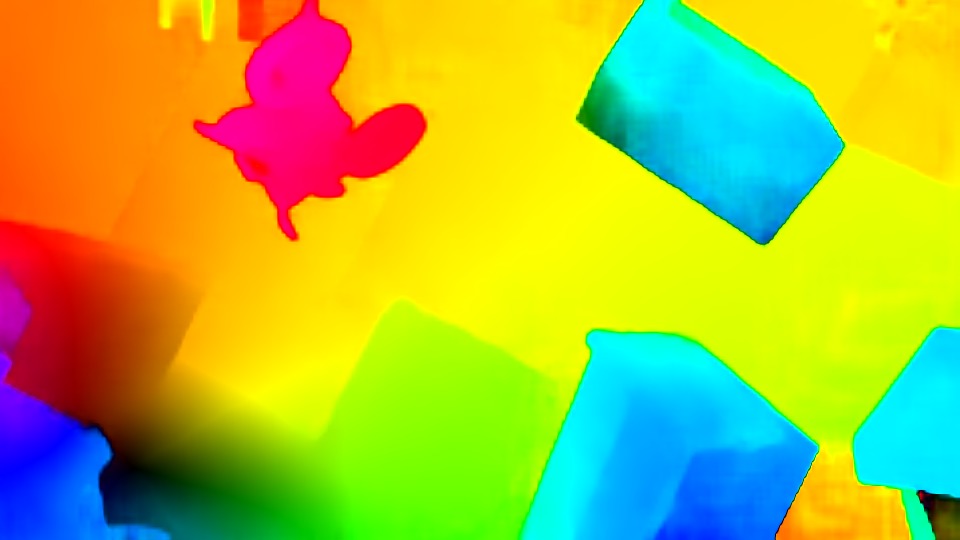} &
        \includegraphics[width=0.22\textwidth]{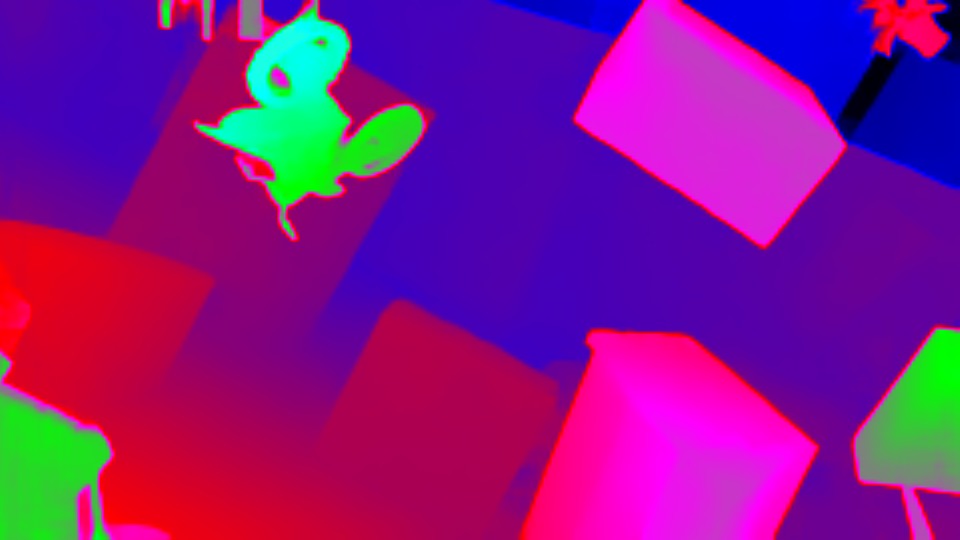} &
        \includegraphics[width=0.22\textwidth]{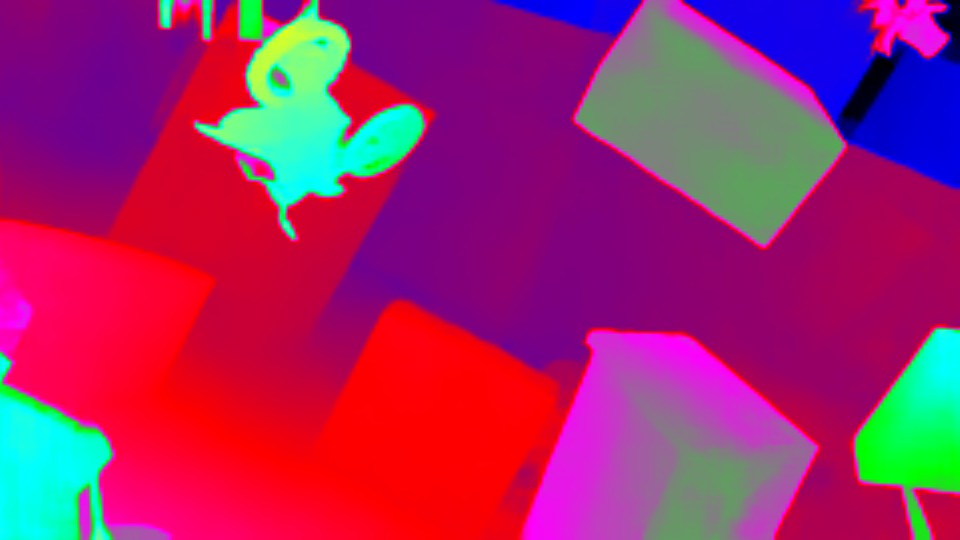} \\ 

        \includegraphics[width=0.22\textwidth]{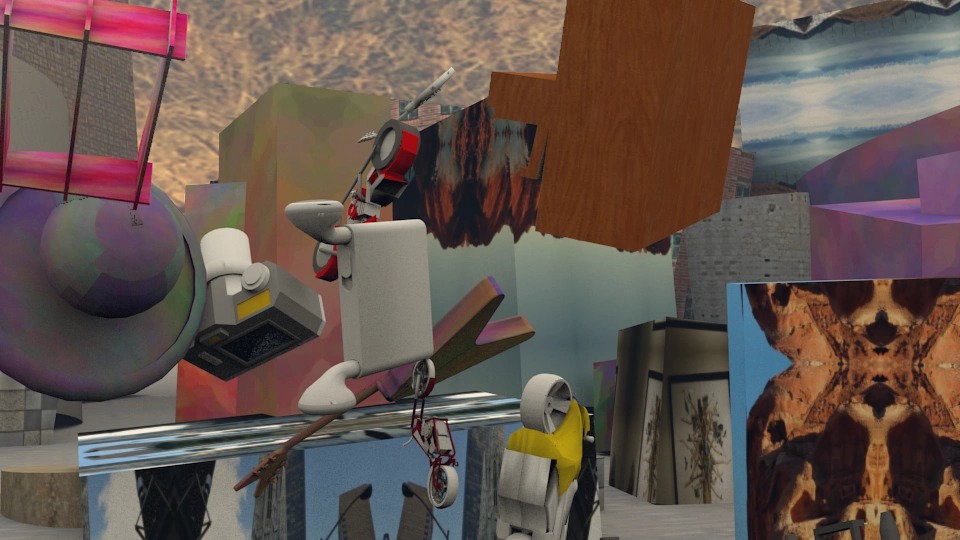} & \includegraphics[width=0.22\textwidth]{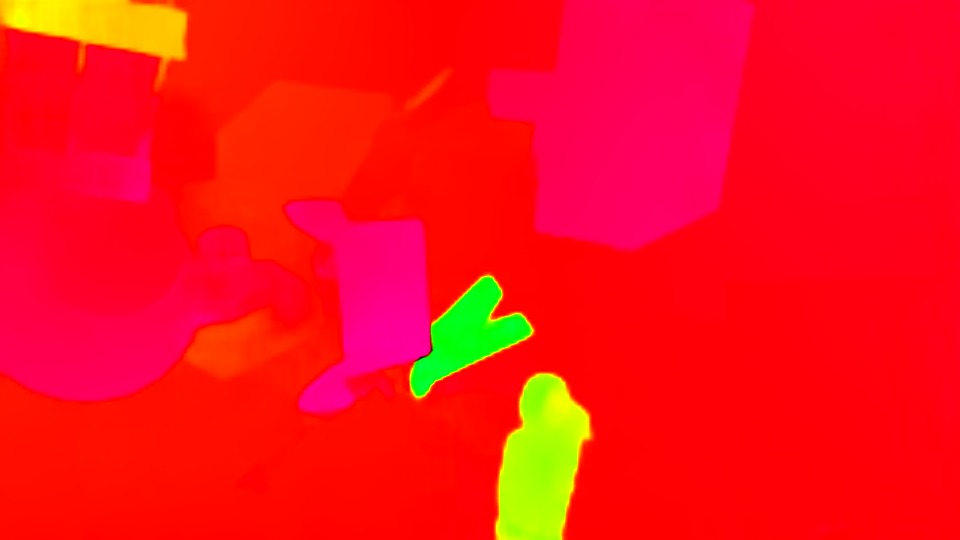} &
        \includegraphics[width=0.22\textwidth]{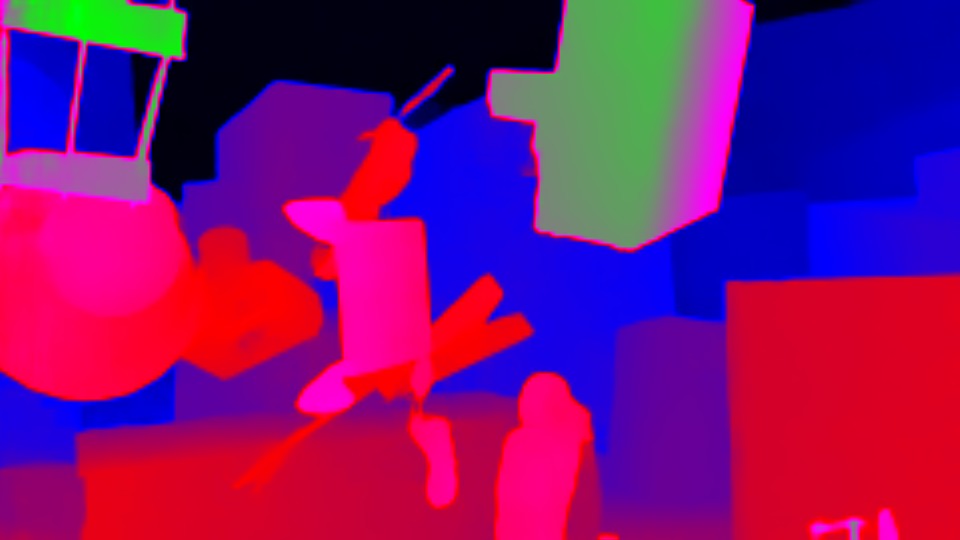} &
        \includegraphics[width=0.22\textwidth]{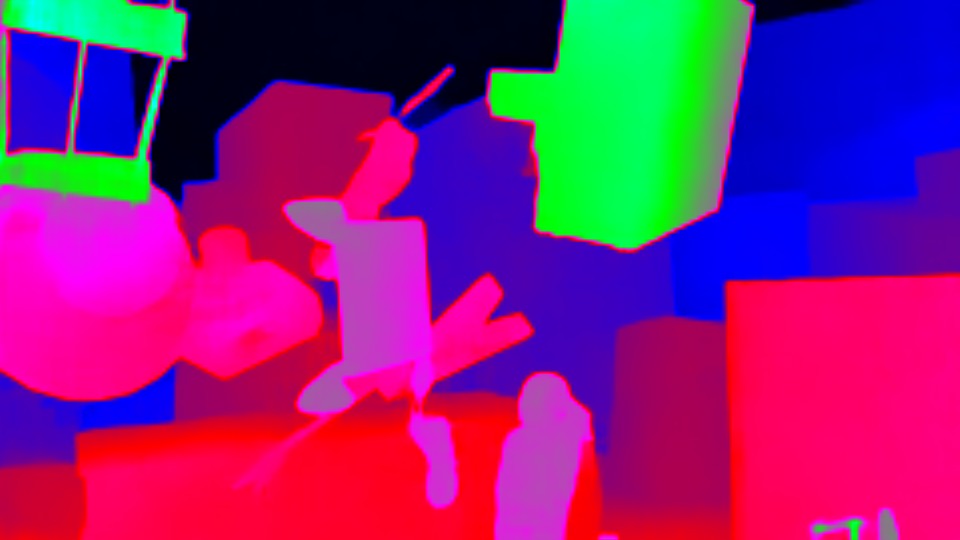} \\

        \includegraphics[width=0.22\textwidth]{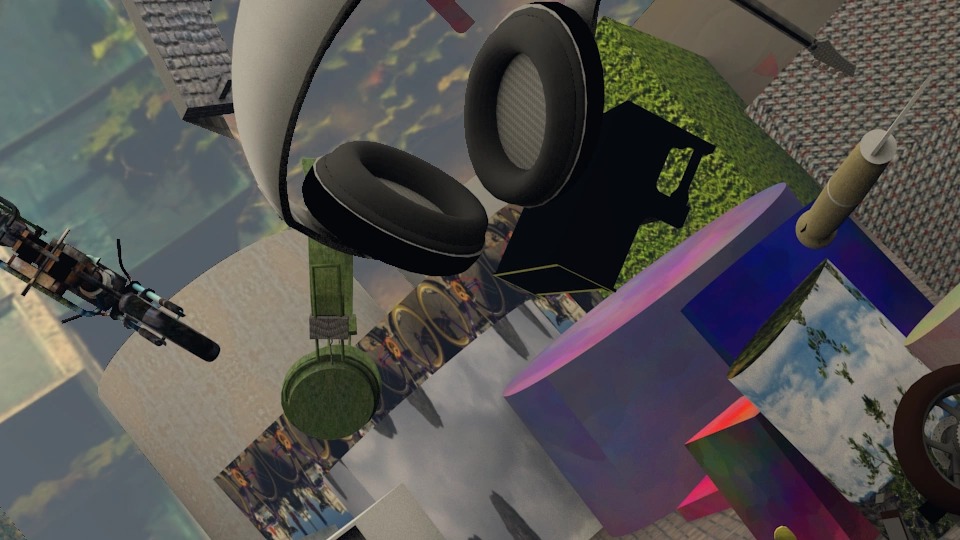} & \includegraphics[width=0.22\textwidth]{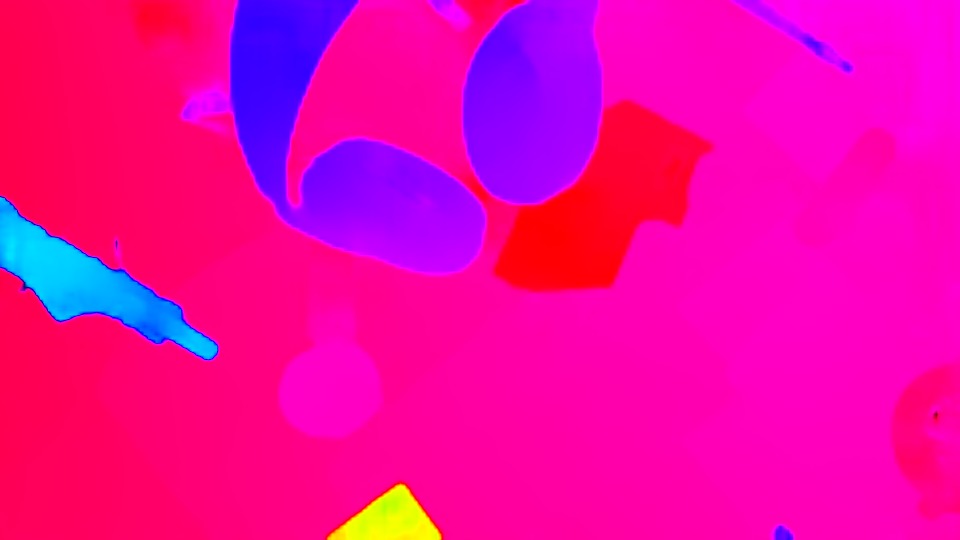} &
        \includegraphics[width=0.22\textwidth]{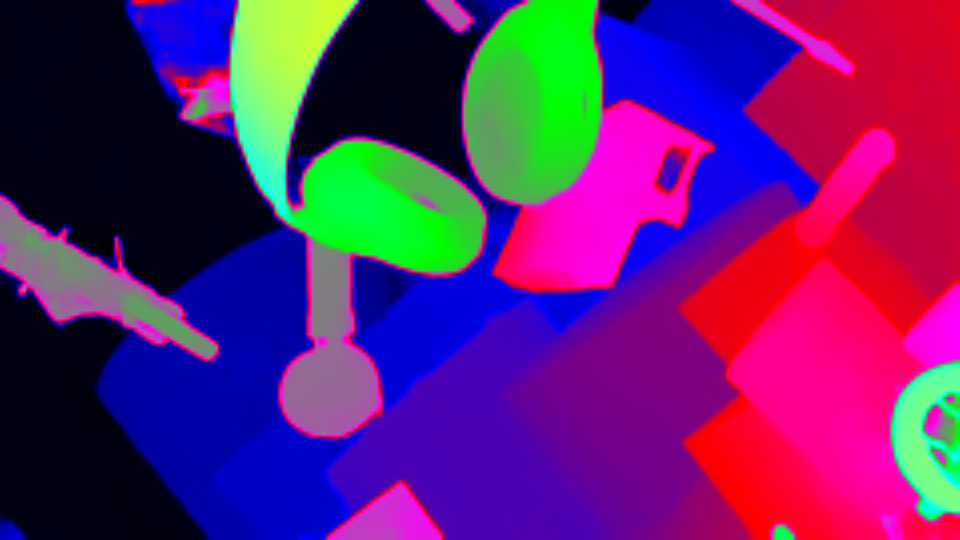} &
        \includegraphics[width=0.22\textwidth]{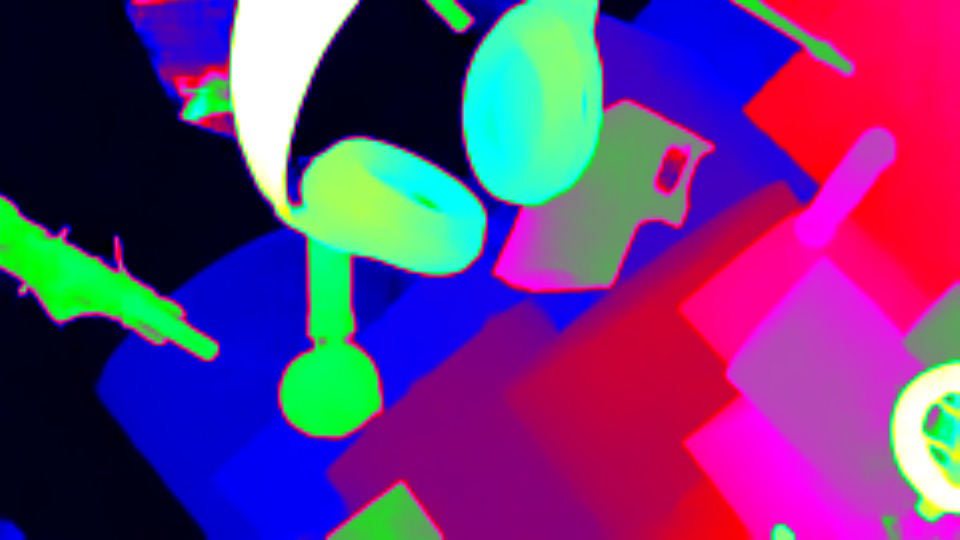} \\

    \end{tabular}
    \caption{\textbf{Qualitative results on FlyingThings 3D \cite{mayer2016large} test split.} From left to right, left image at $t_1$, optical flow, disparity and disparity change.}
    \label{fig:things1}
\end{figure*}

By running a stereo algorithm on stereo pairs at $t_1$ and $t_2$, this will results into different behaviours for star and triangle across the two pairs. Figure \ref{fig:match} shows the cost curves for the two pixels at the different time frames. In this example, we use MC-CNN-acrt \cite{zbontar2016stereo} to generate matching costs. Intuitively, static pixels will show a similar cost distribution across time, as we can notice from the plot on the left where the two curves overlap. In contrast, pixels belonging to moving objects will behave differently, e.g. they will have minimum at different positions since their disparity changes, as shown on the right plot. In particular, in this case the minimum cost changes its position along the disparity axis from 61 to 50, result of the car getting farther from the camera. Anyway, since the neighbouring pixels are almost the same, the two peaks will appear \textit{shifted} by the disparity change. 
As we said in the main paper, the rest of the curve will shrink/enlarge, as occurs for the orange curve being it shrunk with respect to the blue one. 
Of course, difference of appearance introduced by reflections or illumination changes can interfere as well, as we can observe in Figure \ref{fig:match} right, where the curves show different magnitudes in ranges 80-120 despite computed on pixels from the car.

We leverage on correlation scores between $t_1$ and $t_2$ to measure their similarity. By gradually shifting the second curve in a range $[-r_z, r_z]$, we correlate it with the first one to obtain a correlation curve. Figure \ref{fig:corrs} shows correlations scores for static (left) and dynamic (right) pixels taken from Figure \ref{fig:star}. Confirming our hypothesis, static pixels have peaks of correlations on 0 because their cost curve are almost overlapped across time. About dynamic pixels, we can see a peak in correspondence of -11, i.e. the change in disparity between frame $t_1$ and $t_2$. Thus, by designing our novel 3D correlation layer, we are able to find correspondences between the cost curves by correlating in a 2D search window in the image plus a further 1D search range on the disparity change dimension.
To keep computational cost affordable, we limit the 3D search range to $9\times9\times1$. Nevertheless, according to our experiments in the main paper, it results enough to improve the accuracy on the three estimations, as the network learns to distinguish the different behaviors of the correlation curves for static and moving pixels.

\begin{figure*}[h]
    \centering
    \renewcommand{\tabcolsep}{1pt}
    \begin{tabular}{cccc}

        \includegraphics[width=0.22\textwidth]{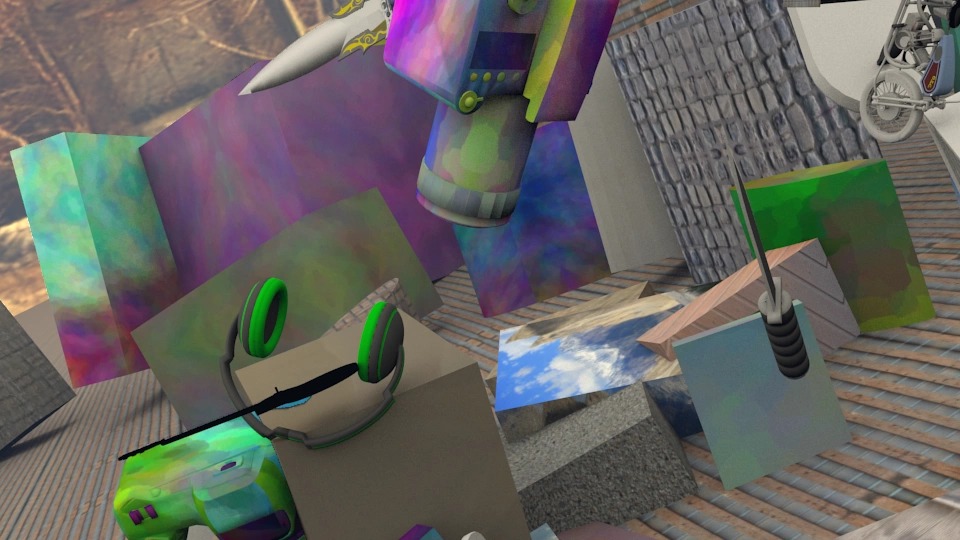} & \includegraphics[width=0.22\textwidth]{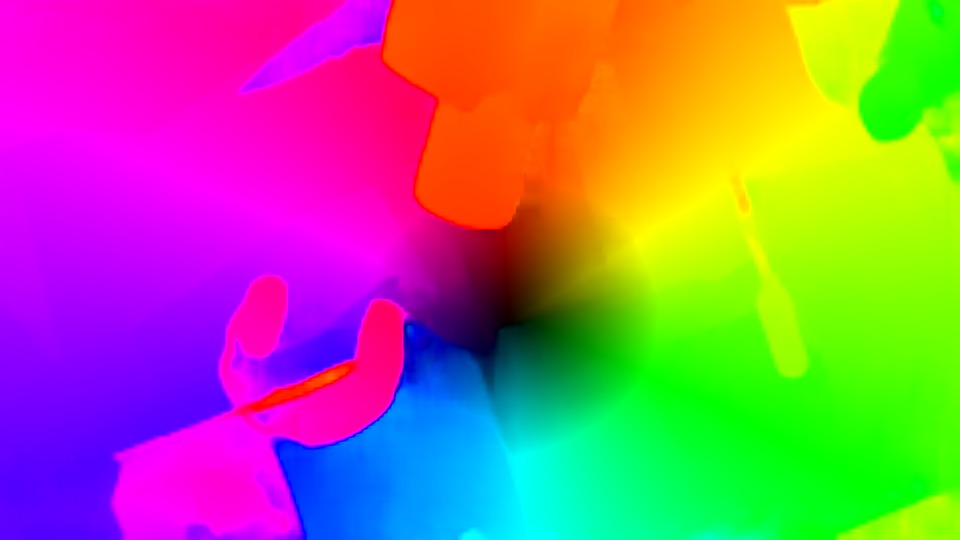} &
        \includegraphics[width=0.22\textwidth]{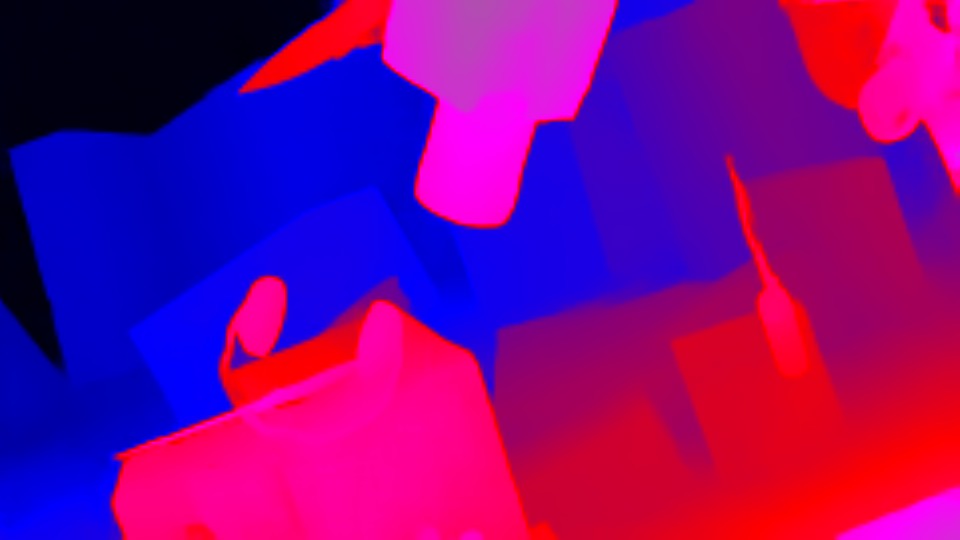} &
        \includegraphics[width=0.22\textwidth]{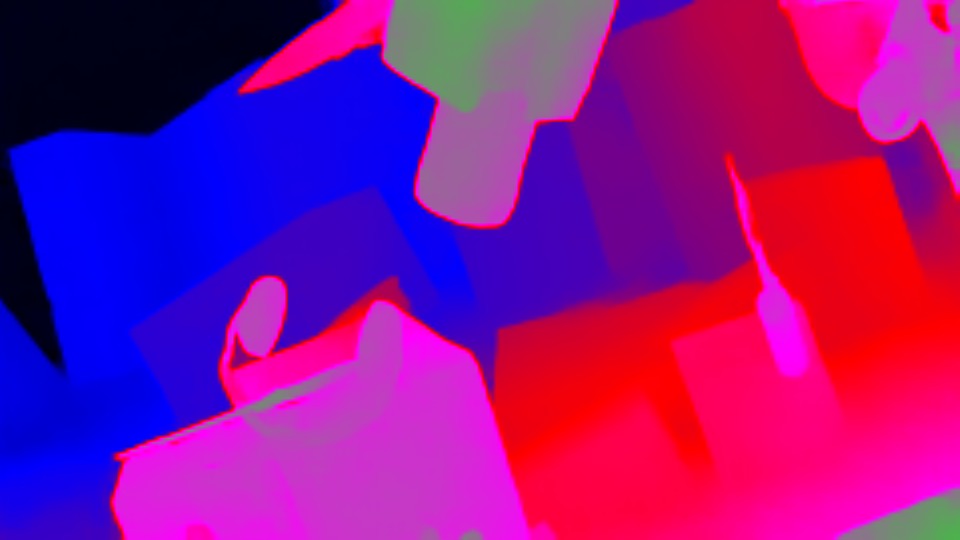} \\
        
        \includegraphics[width=0.22\textwidth]{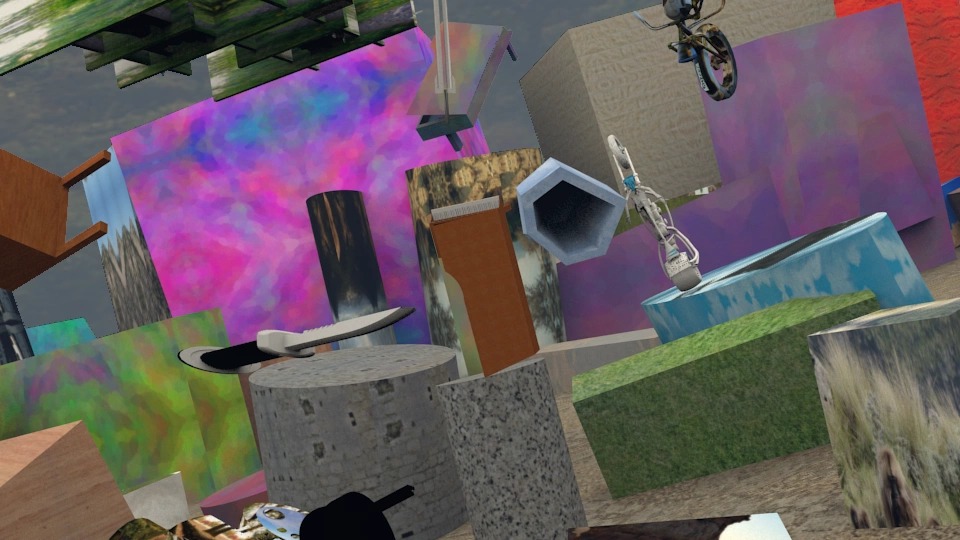} & \includegraphics[width=0.22\textwidth]{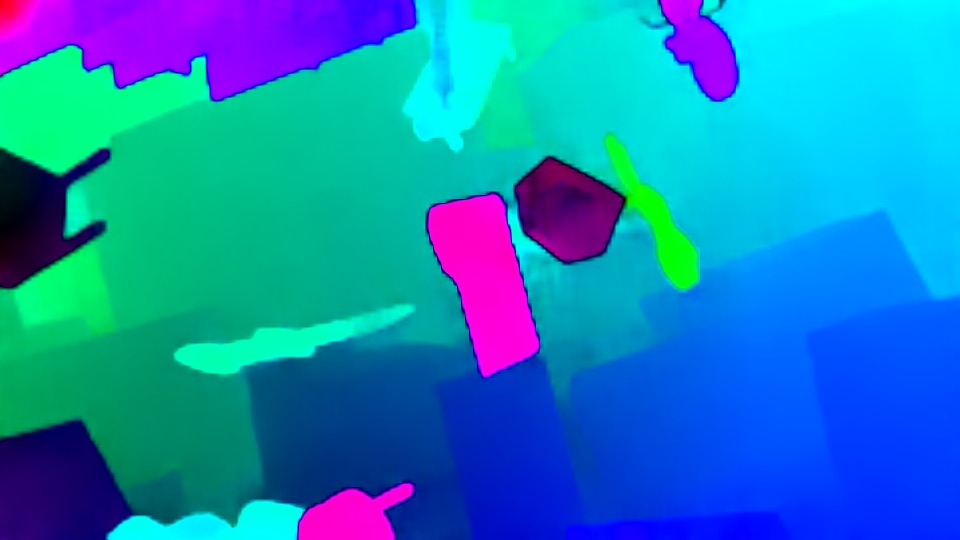} &
        \includegraphics[width=0.22\textwidth]{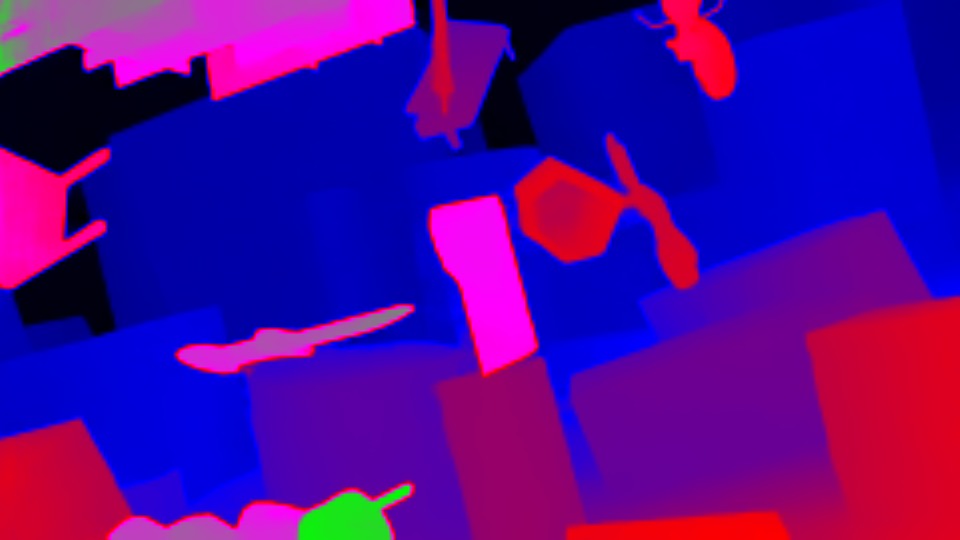} &
        \includegraphics[width=0.22\textwidth]{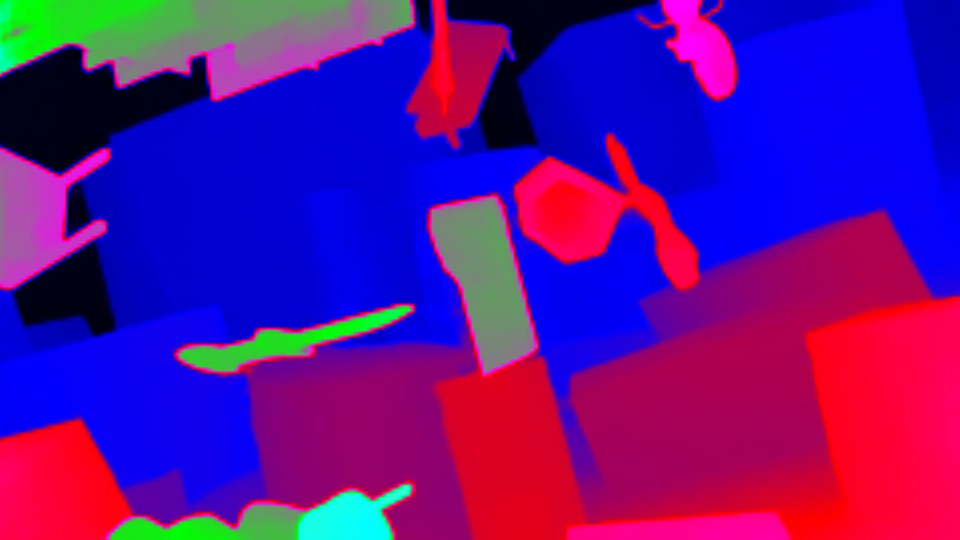} \\                

        \includegraphics[width=0.22\textwidth]{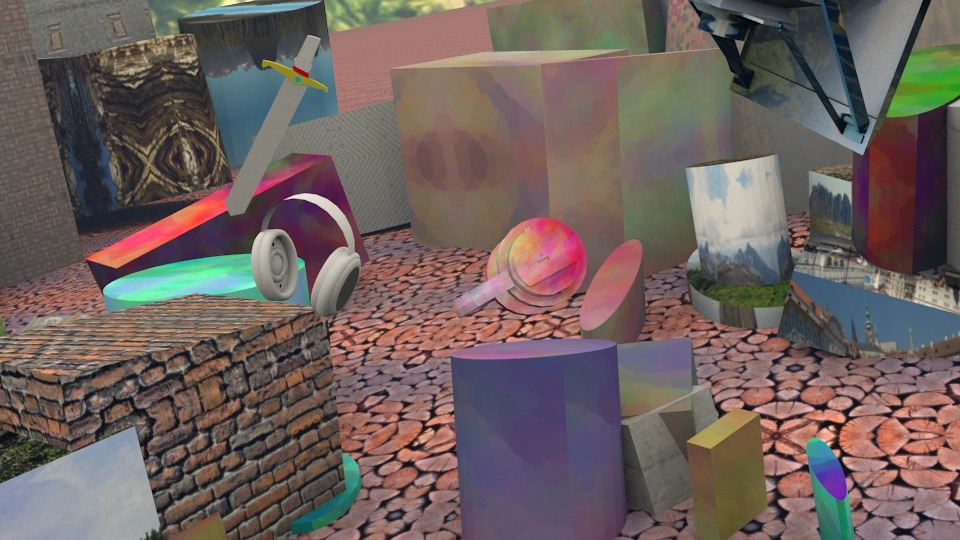} & \includegraphics[width=0.22\textwidth]{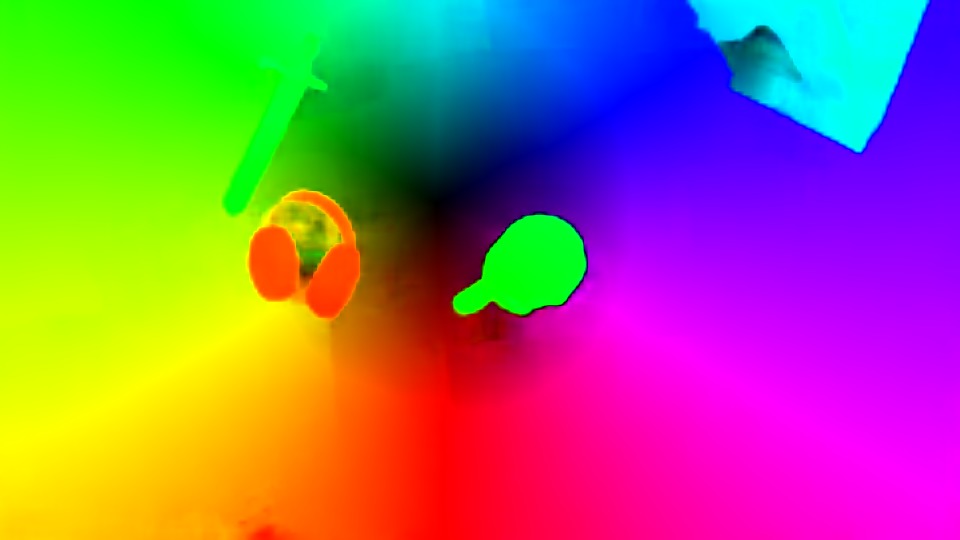} &
        \includegraphics[width=0.22\textwidth]{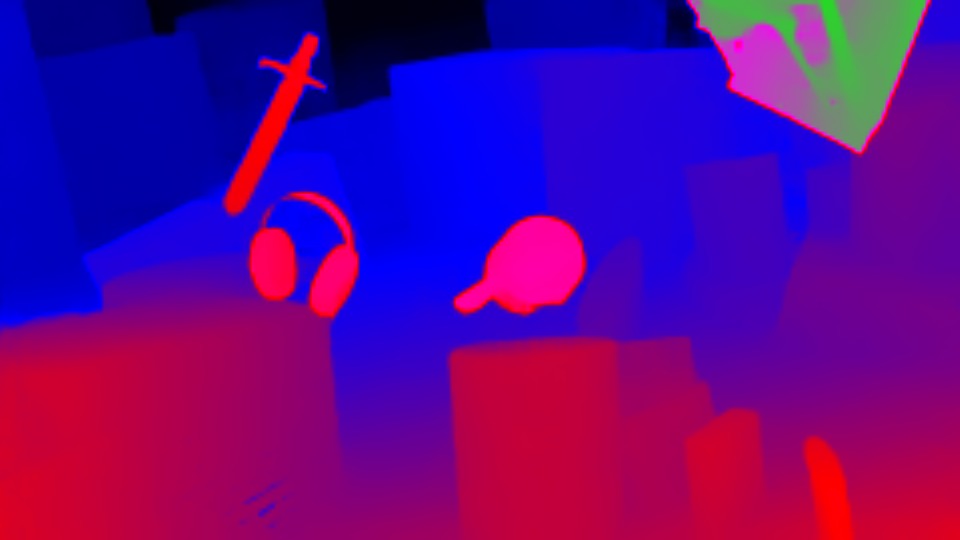} &
        \includegraphics[width=0.22\textwidth]{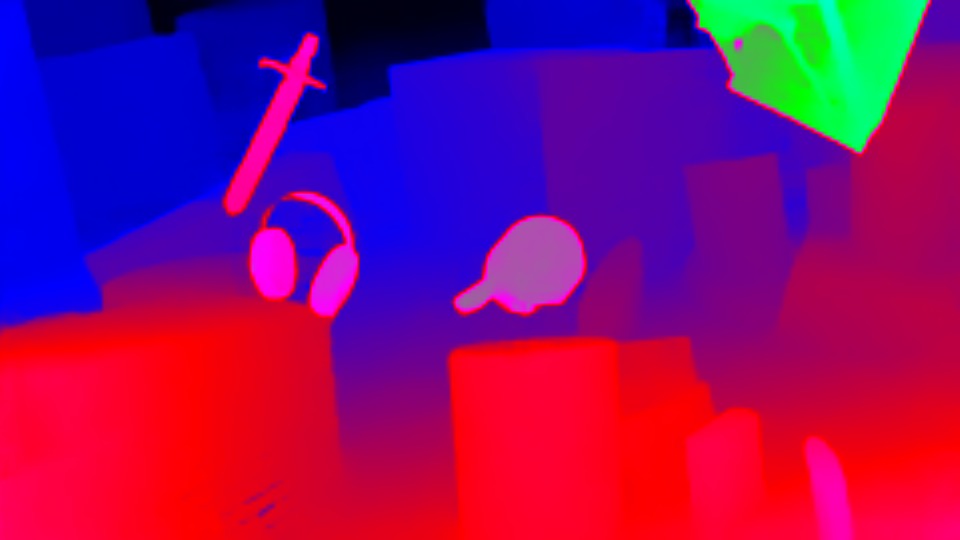} \\

        \includegraphics[width=0.22\textwidth]{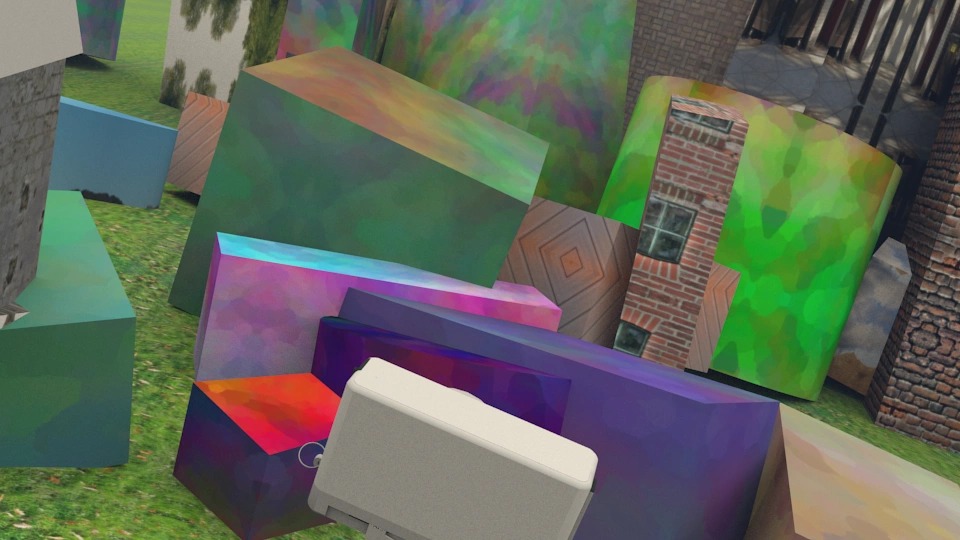} & \includegraphics[width=0.22\textwidth]{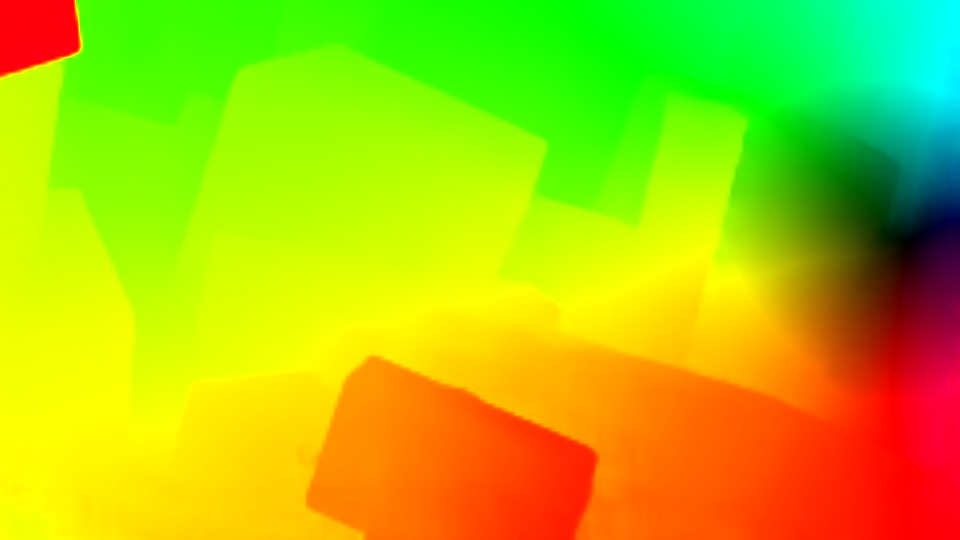} &
        \includegraphics[width=0.22\textwidth]{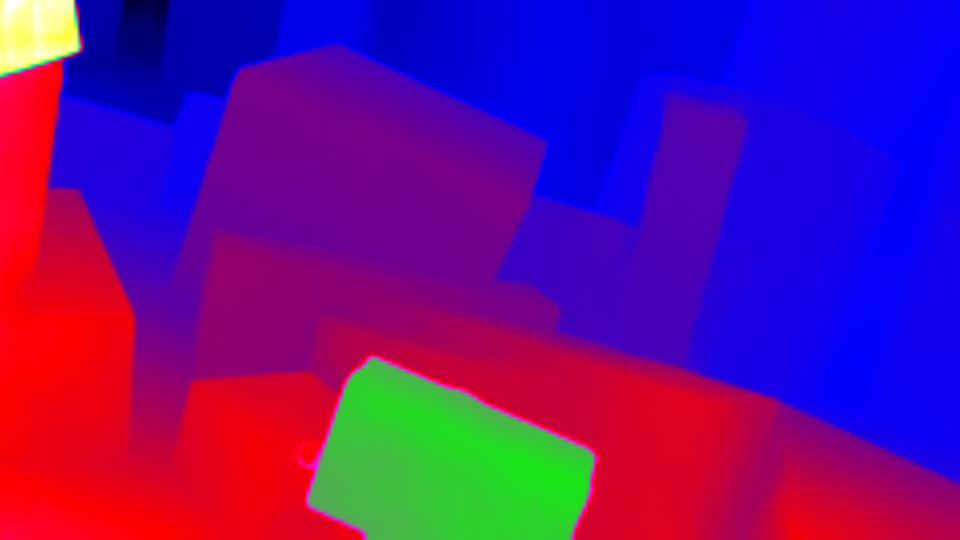} &
        \includegraphics[width=0.22\textwidth]{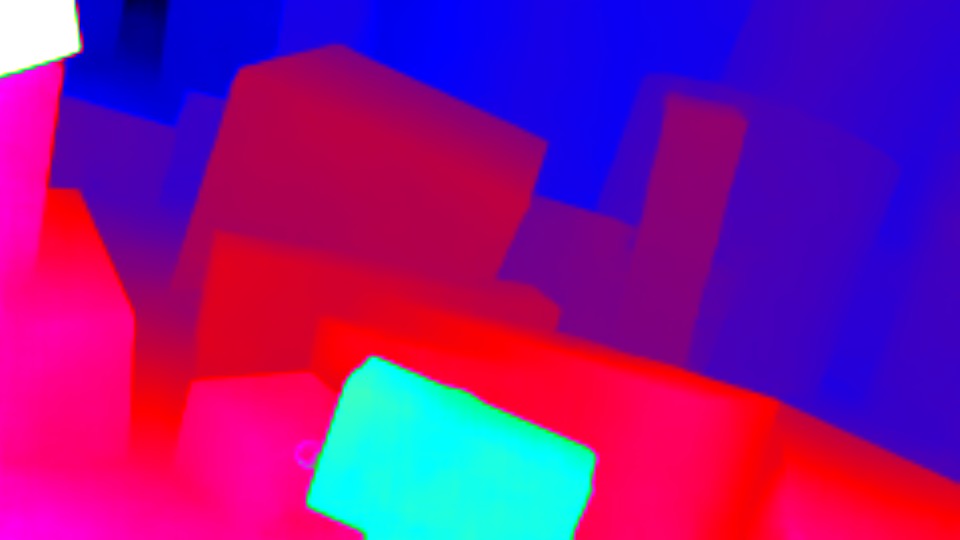} \\

        \includegraphics[width=0.22\textwidth]{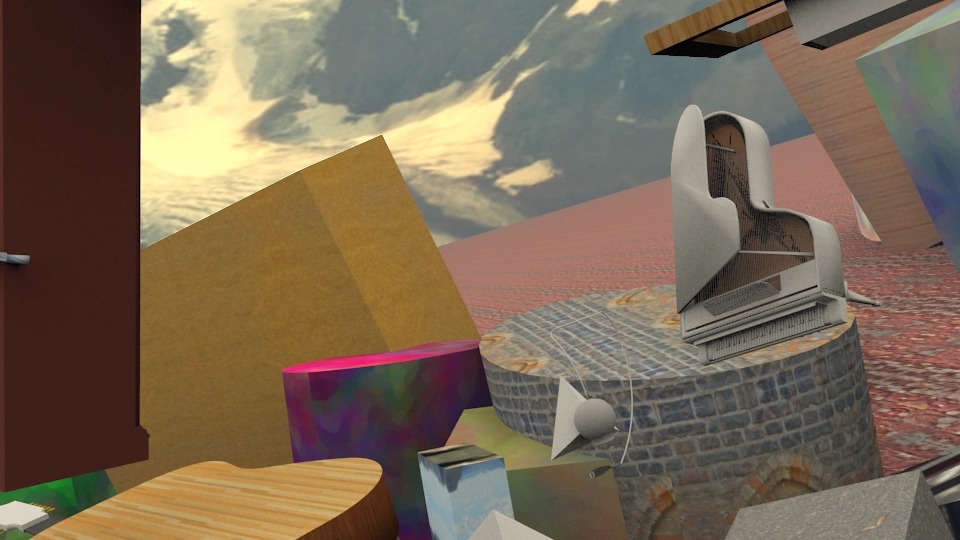} & \includegraphics[width=0.22\textwidth]{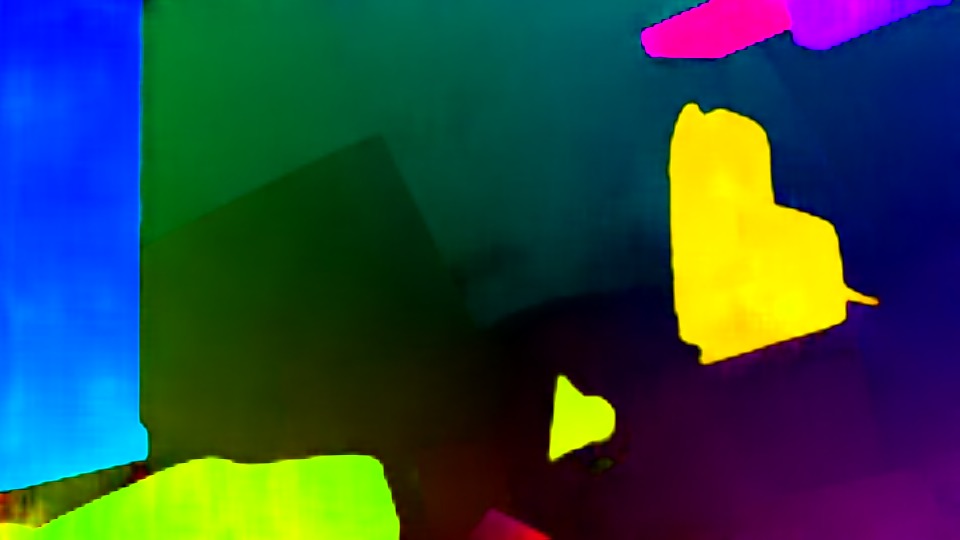} &
        \includegraphics[width=0.22\textwidth]{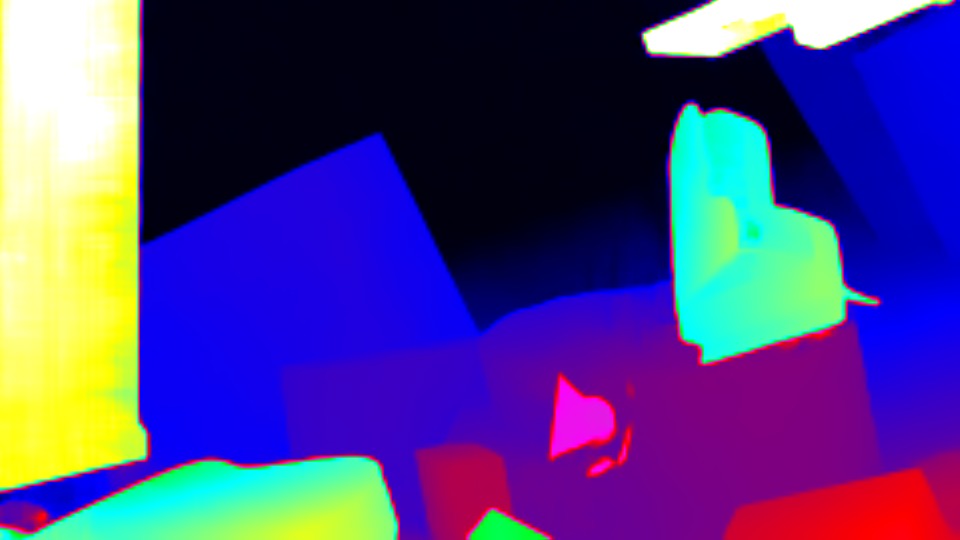} &
        \includegraphics[width=0.22\textwidth]{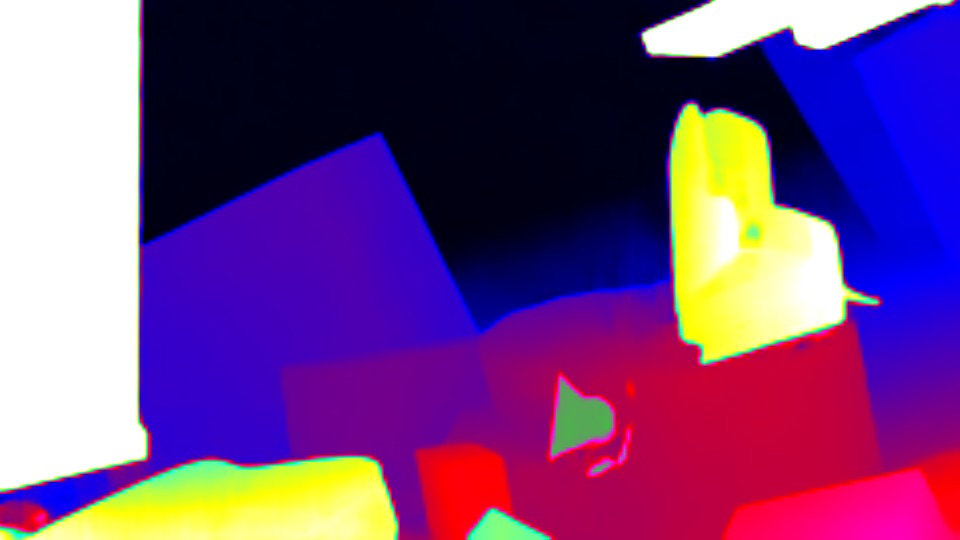} \\

        \includegraphics[width=0.22\textwidth]{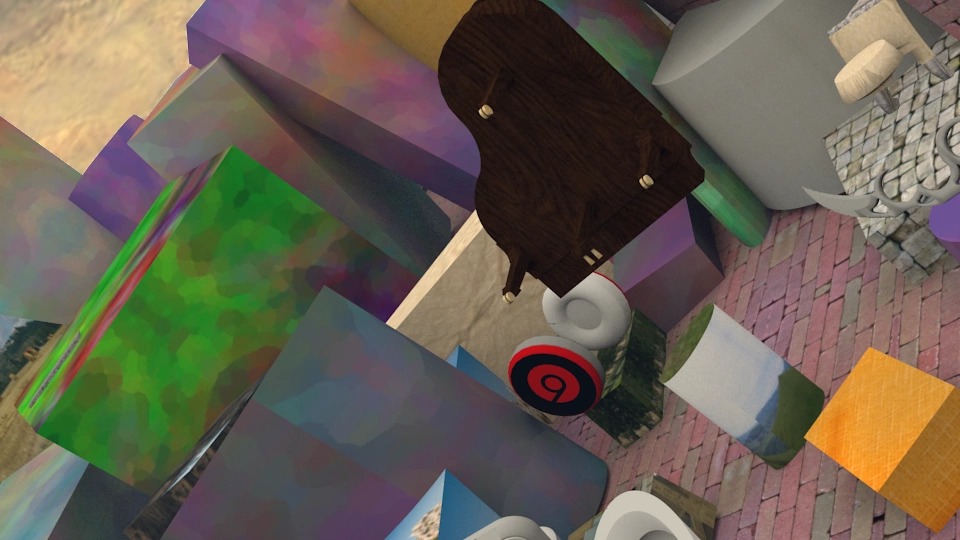} & \includegraphics[width=0.22\textwidth]{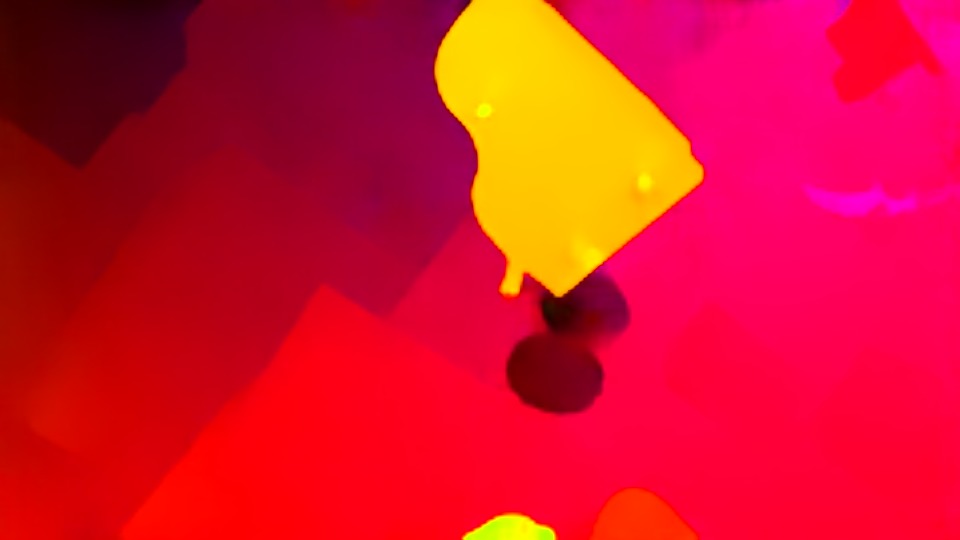} &
        \includegraphics[width=0.22\textwidth]{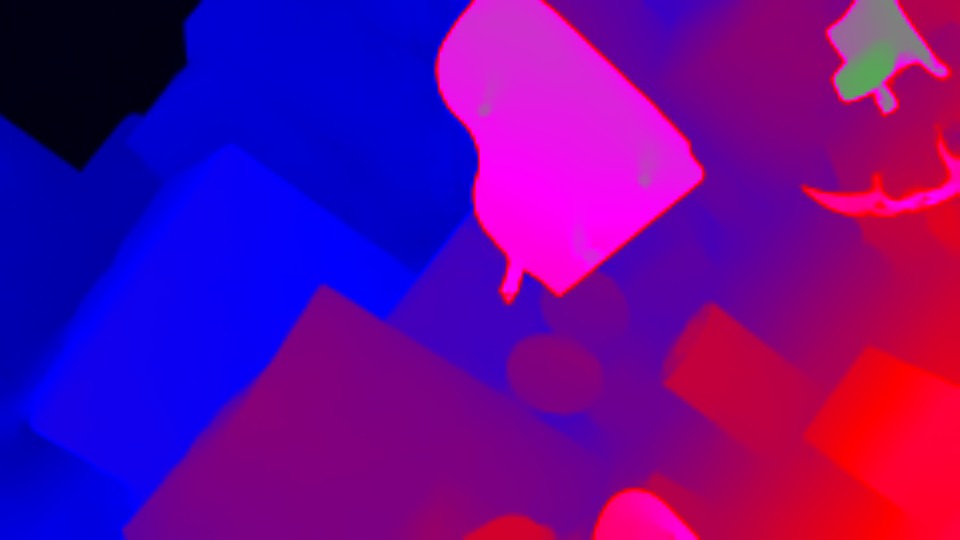} &
        \includegraphics[width=0.22\textwidth]{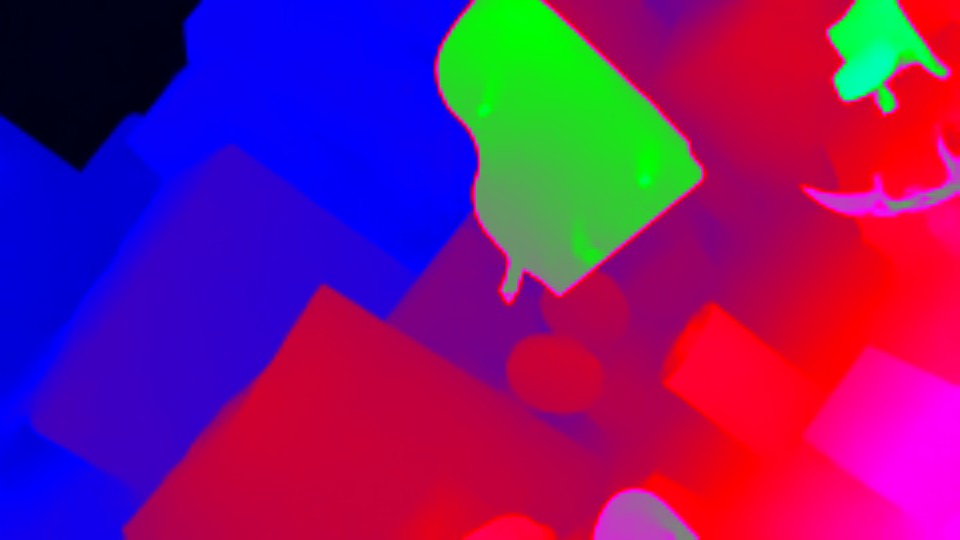} \\
       
    \end{tabular}
    \caption{\textbf{Qualitative esults on FlyingThings 3D \cite{mayer2016large} test split.} From left to right, left image at $t_1$, optical flow, disparity and disparity change.}
    \label{fig:things2}
\end{figure*}

\begin{figure*}
    \centering
    \renewcommand{\tabcolsep}{1pt}
    \begin{tabular}{cccc}
        \includegraphics[width=0.24\textwidth]{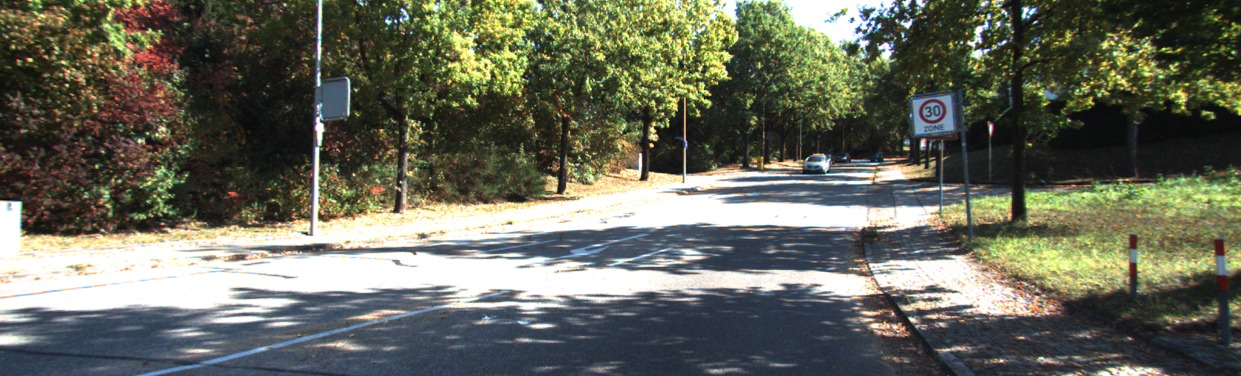} & \includegraphics[width=0.24\textwidth]{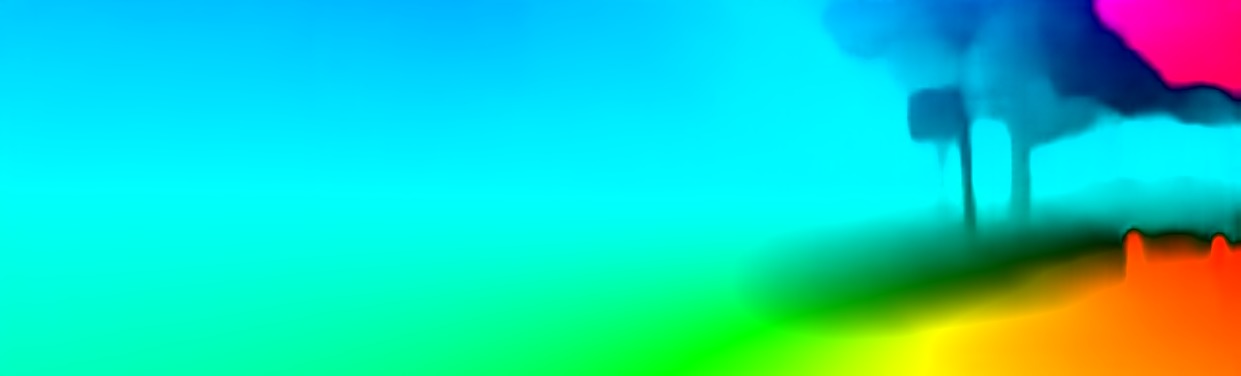} &
        \includegraphics[width=0.24\textwidth]{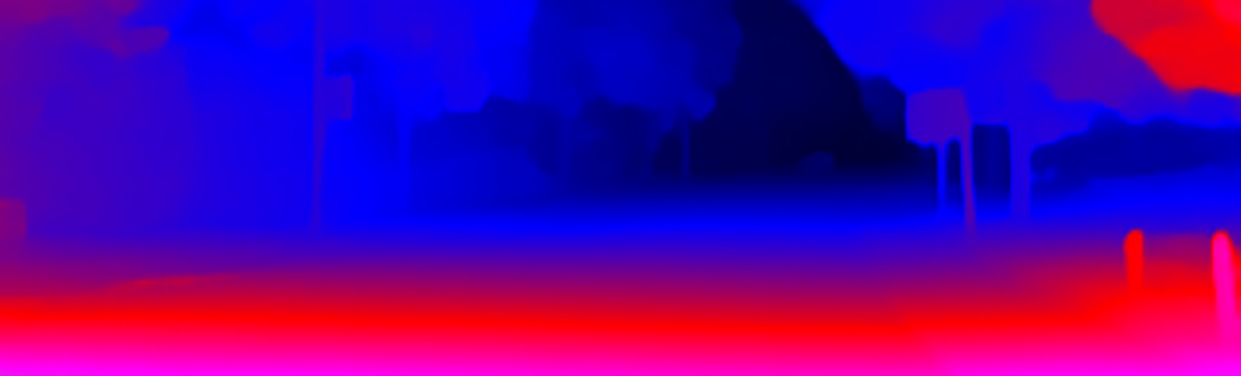} &
        \includegraphics[width=0.24\textwidth]{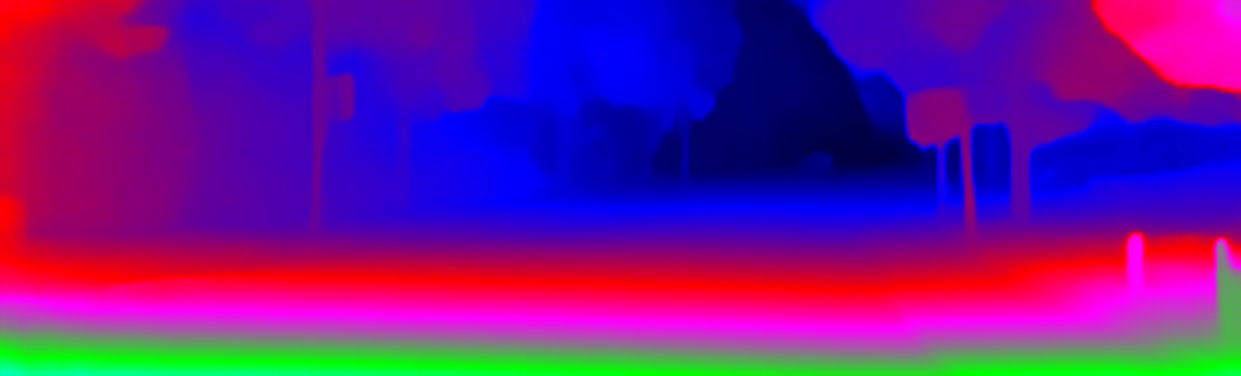} \\ 
        
        \includegraphics[width=0.24\textwidth]{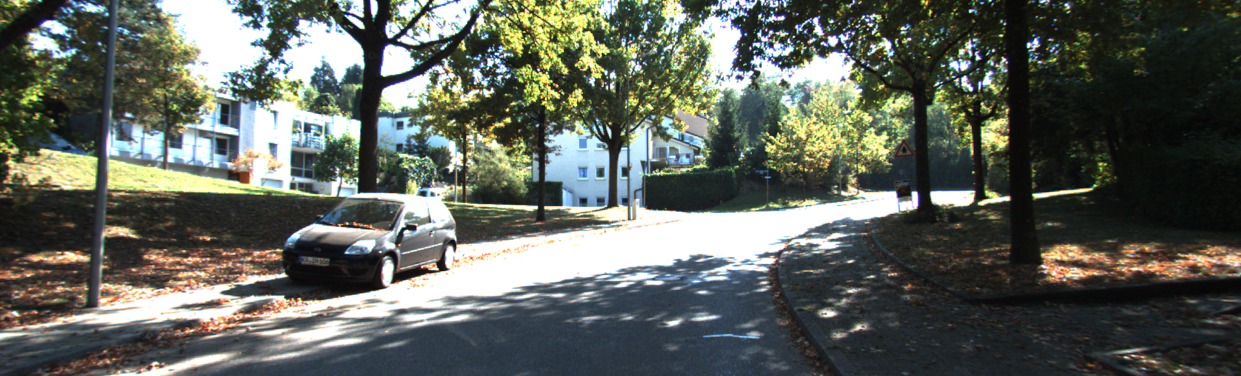} & \includegraphics[width=0.24\textwidth]{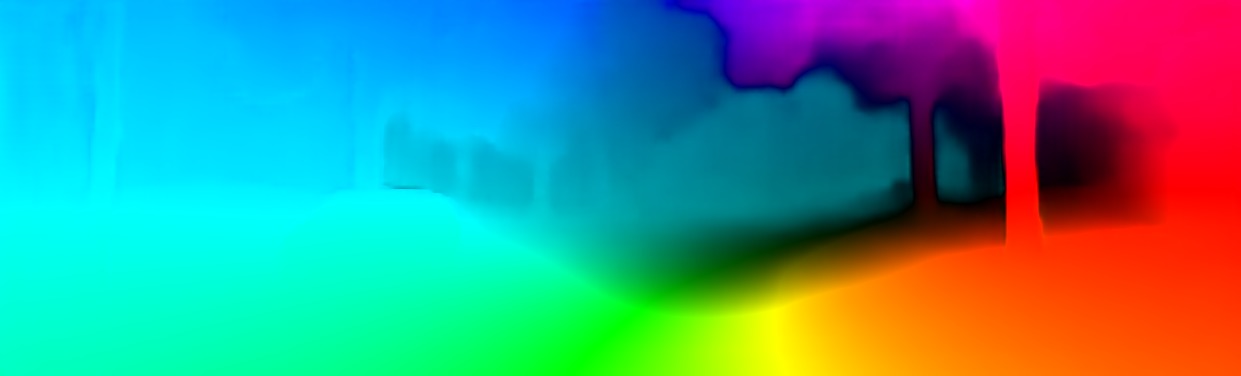} &
        \includegraphics[width=0.24\textwidth]{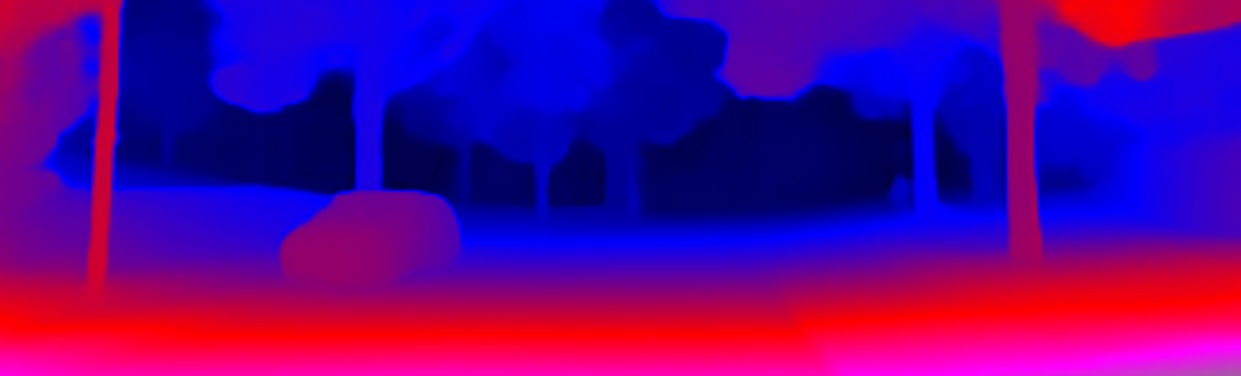} &
        \includegraphics[width=0.24\textwidth]{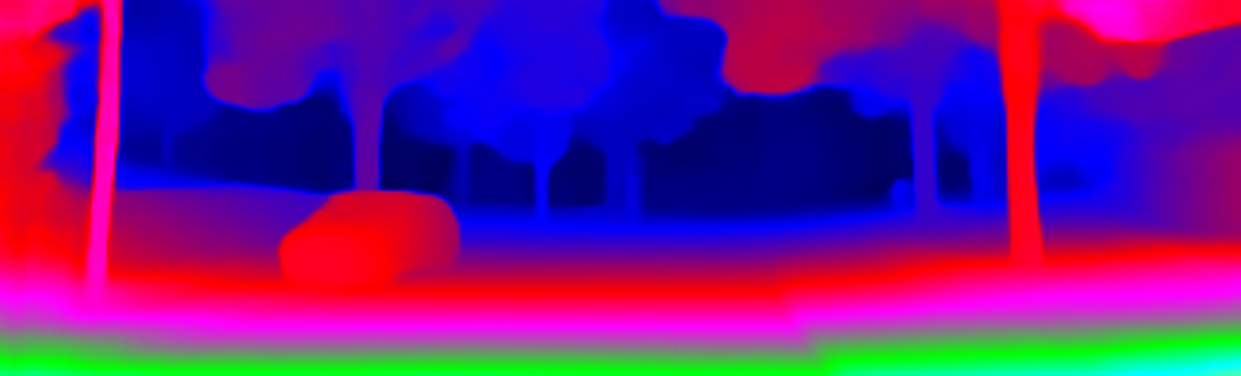} \\     

        \includegraphics[width=0.24\textwidth]{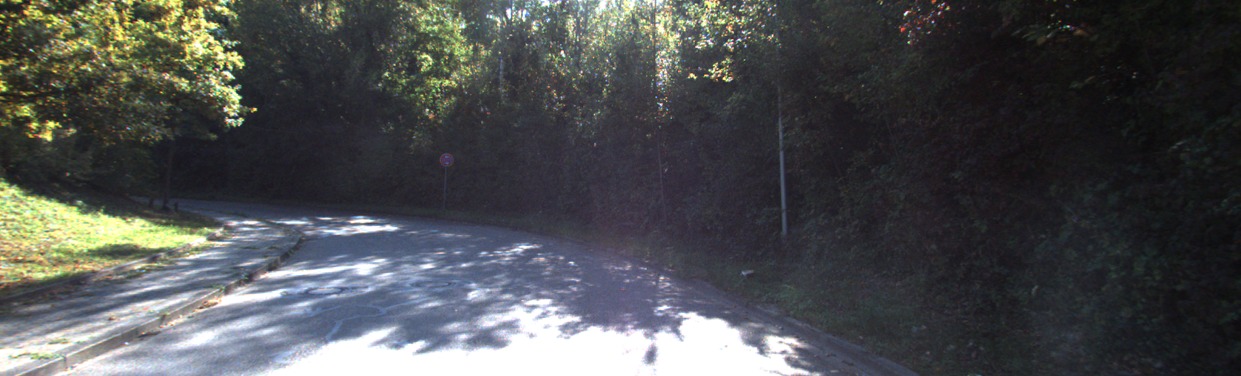} & \includegraphics[width=0.24\textwidth]{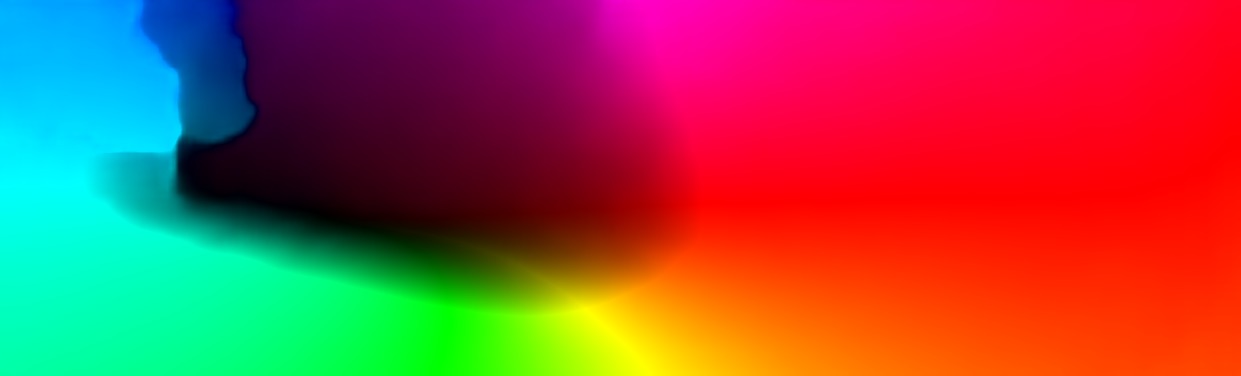} &
        \includegraphics[width=0.24\textwidth]{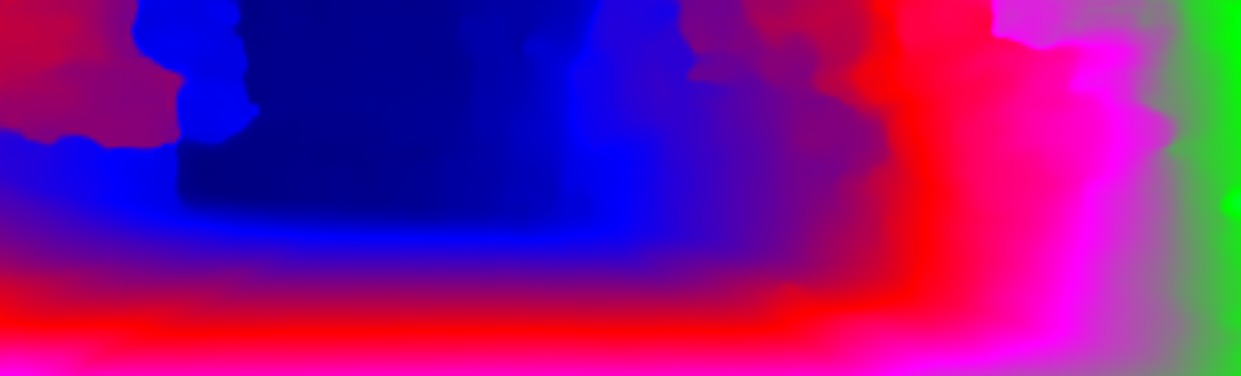} &
        \includegraphics[width=0.24\textwidth]{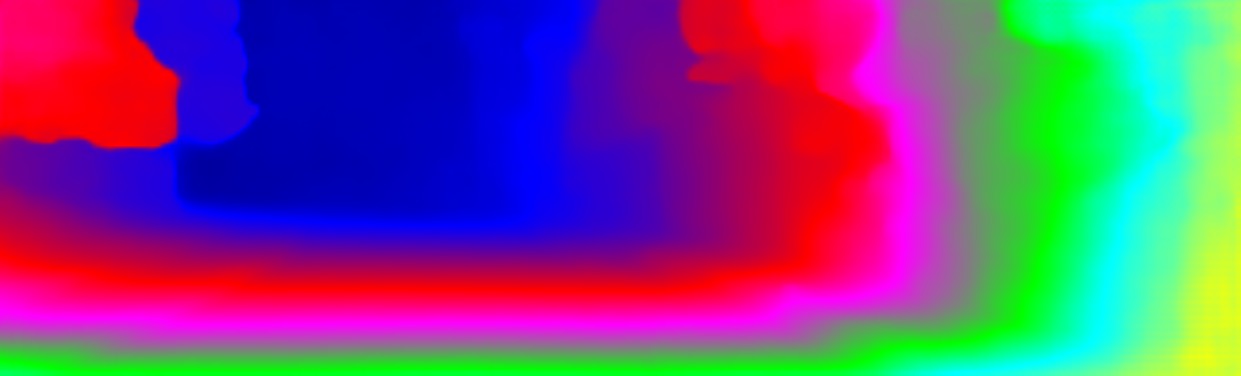} \\ 

        \includegraphics[width=0.24\textwidth]{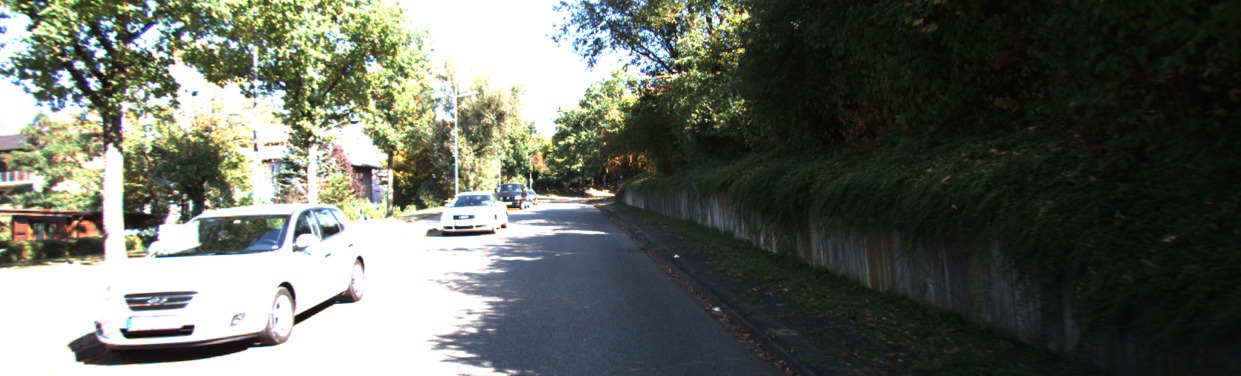} & \includegraphics[width=0.24\textwidth]{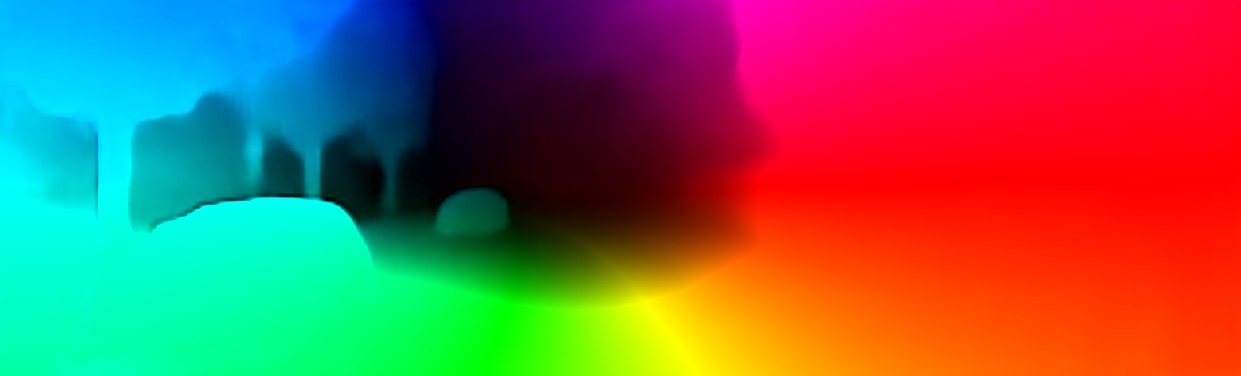} &
        \includegraphics[width=0.24\textwidth]{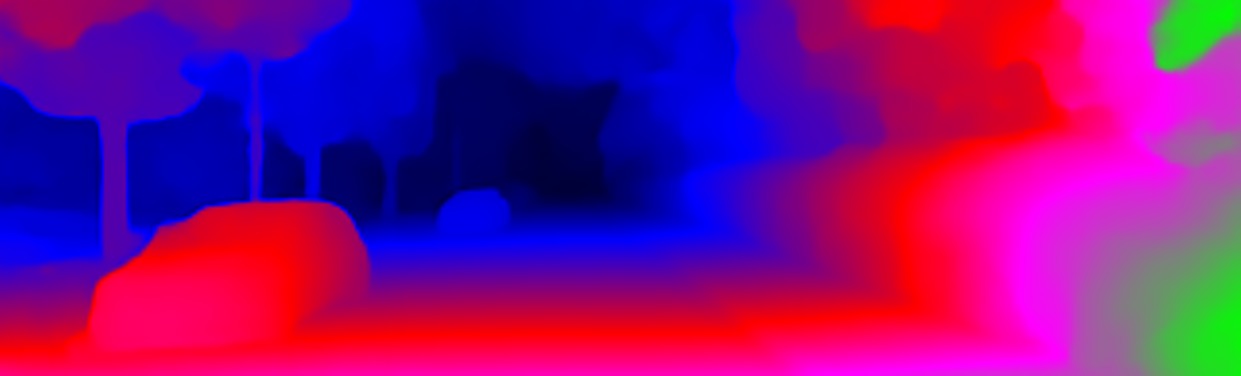} &
        \includegraphics[width=0.24\textwidth]{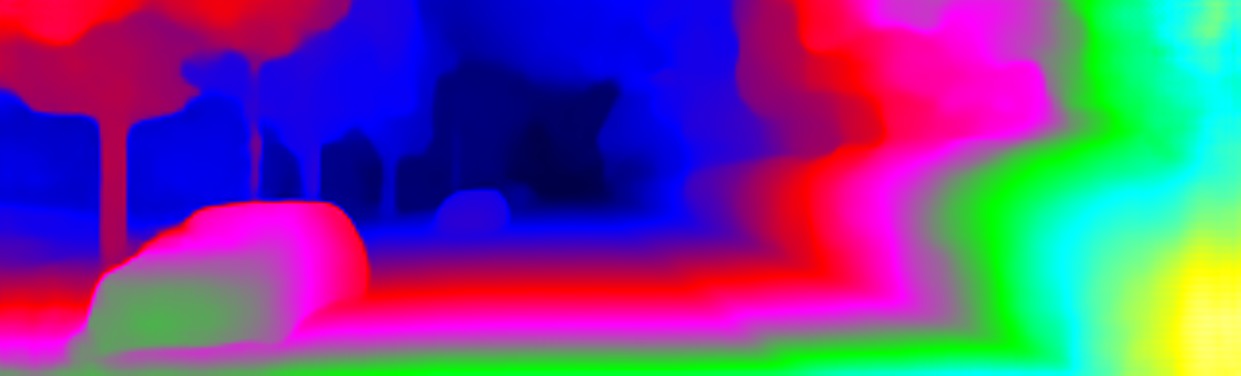} \\ 

        \includegraphics[width=0.24\textwidth]{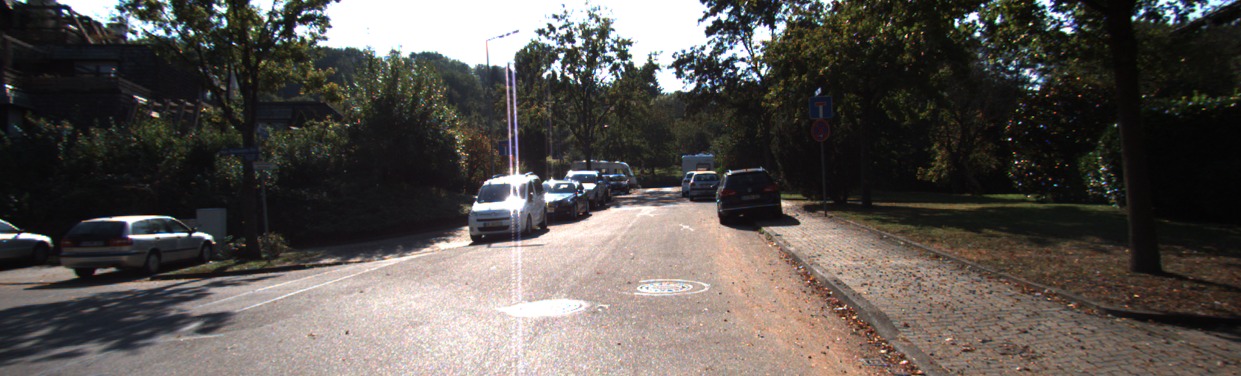} & \includegraphics[width=0.24\textwidth]{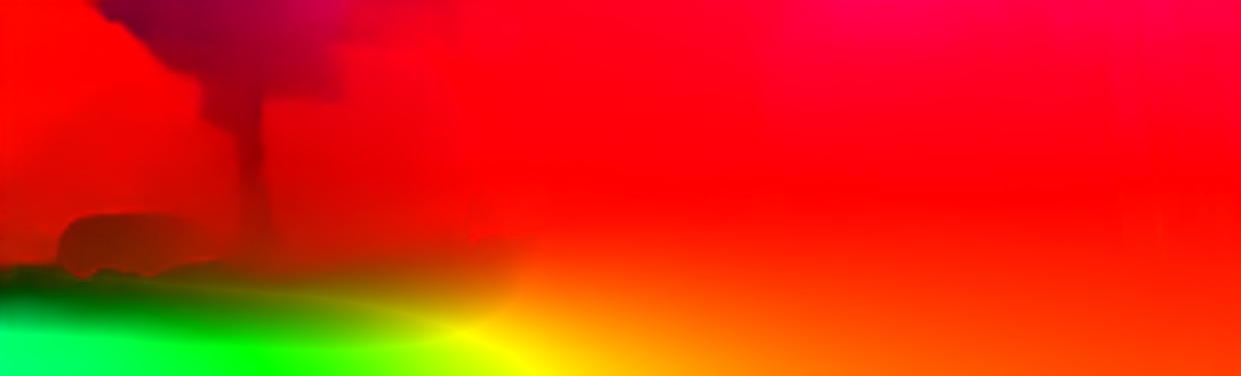} &
        \includegraphics[width=0.24\textwidth]{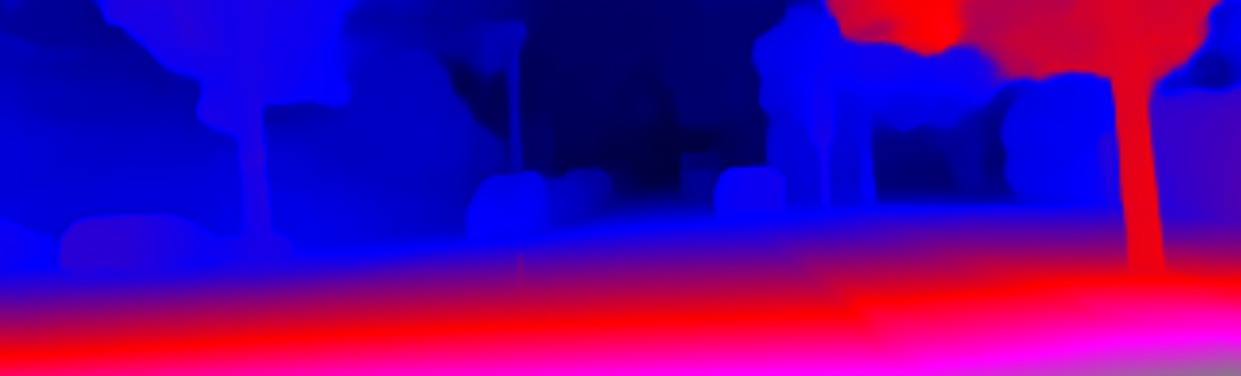} &
        \includegraphics[width=0.24\textwidth]{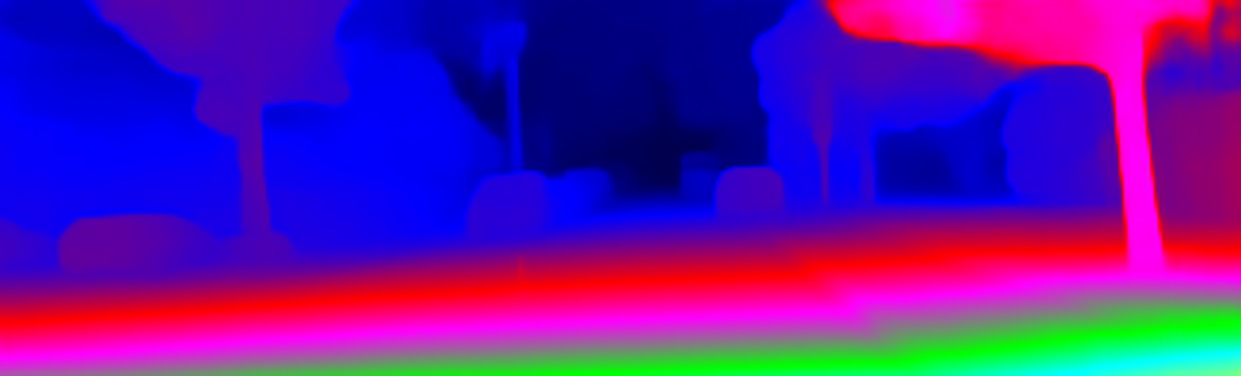} \\   

        \includegraphics[width=0.24\textwidth]{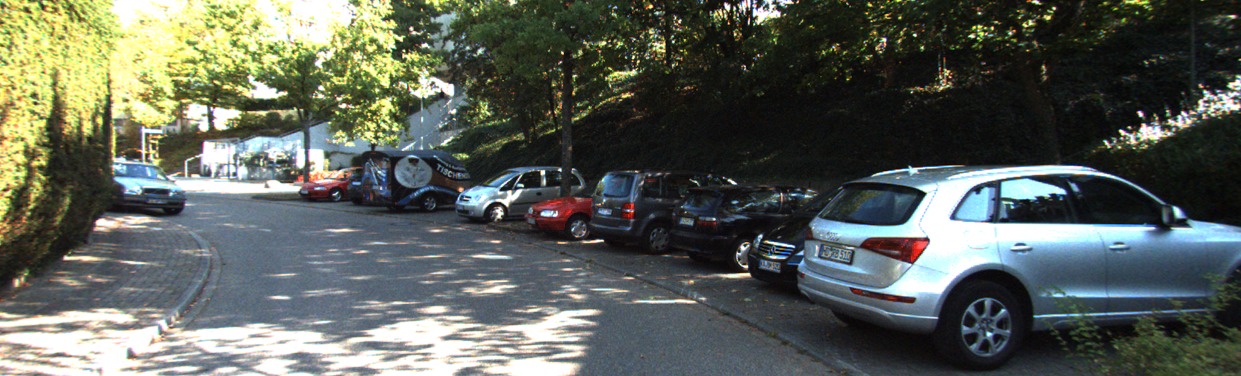} & \includegraphics[width=0.24\textwidth]{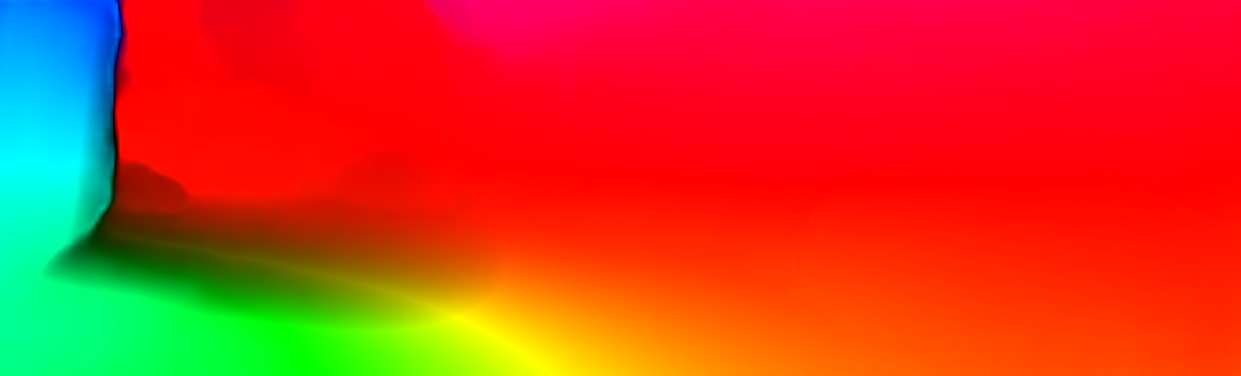} &
        \includegraphics[width=0.24\textwidth]{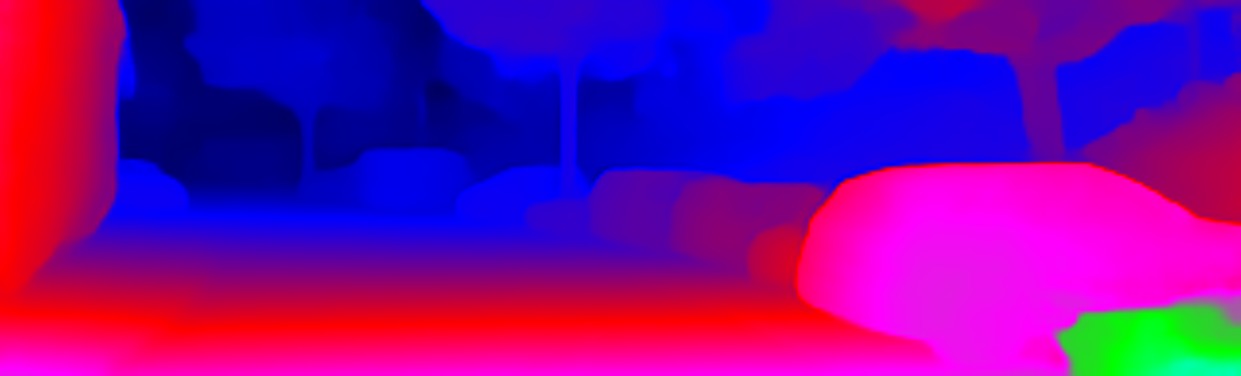} &
        \includegraphics[width=0.24\textwidth]{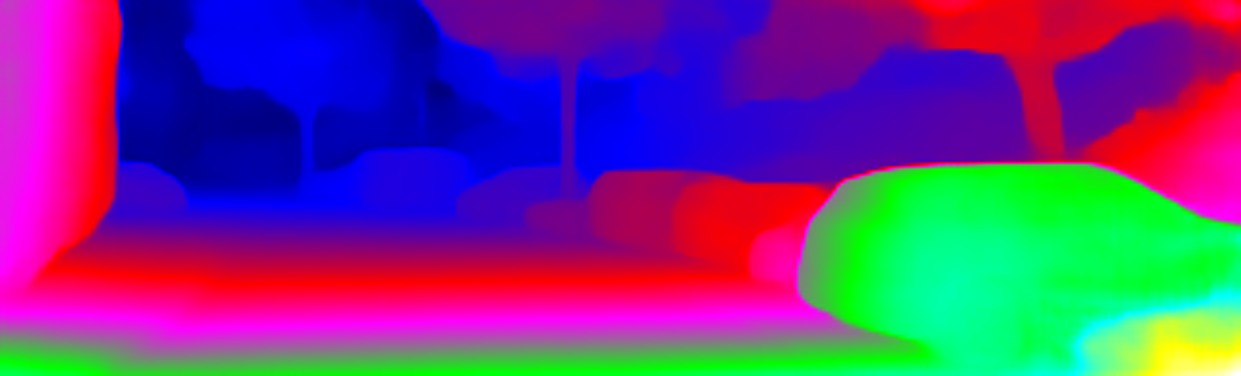} \\

        \includegraphics[width=0.24\textwidth]{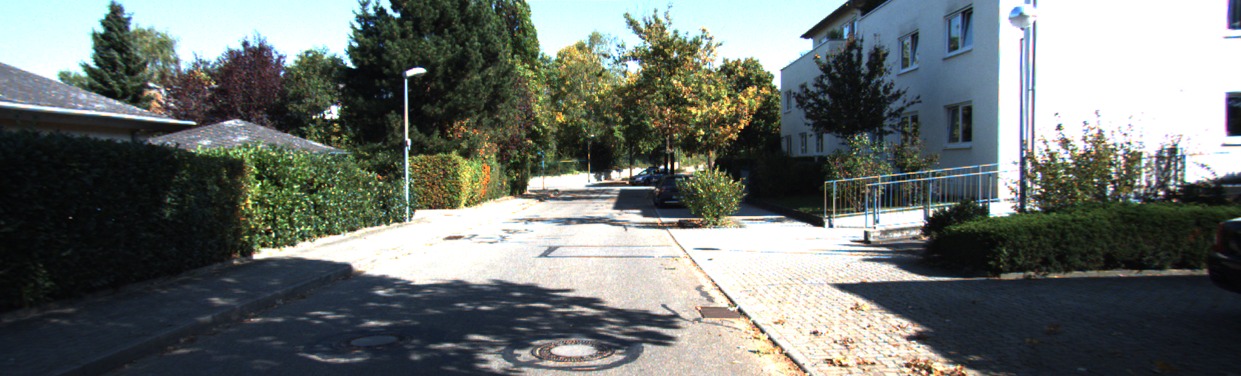} & \includegraphics[width=0.24\textwidth]{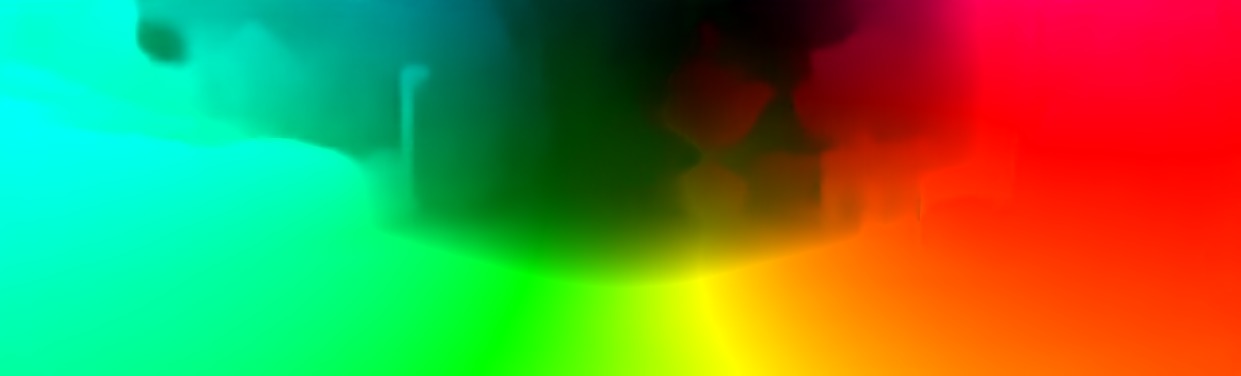} &
        \includegraphics[width=0.24\textwidth]{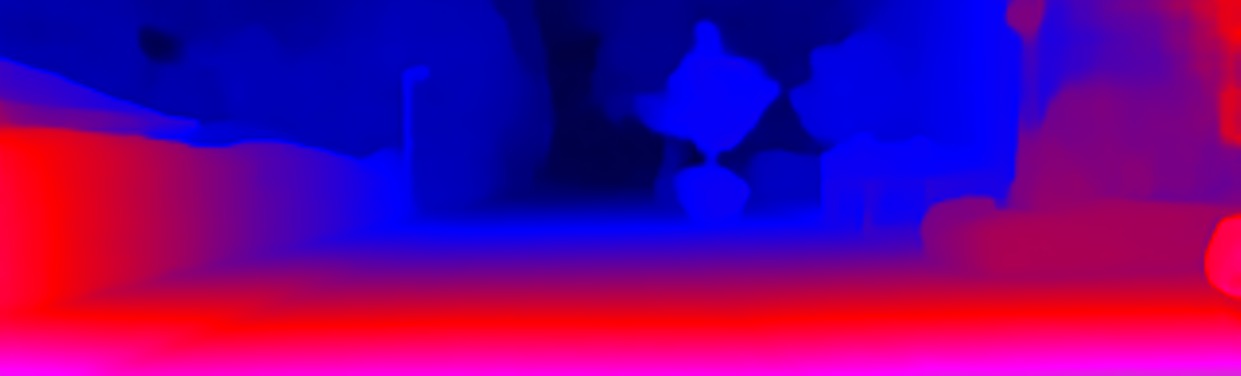} &
        \includegraphics[width=0.24\textwidth]{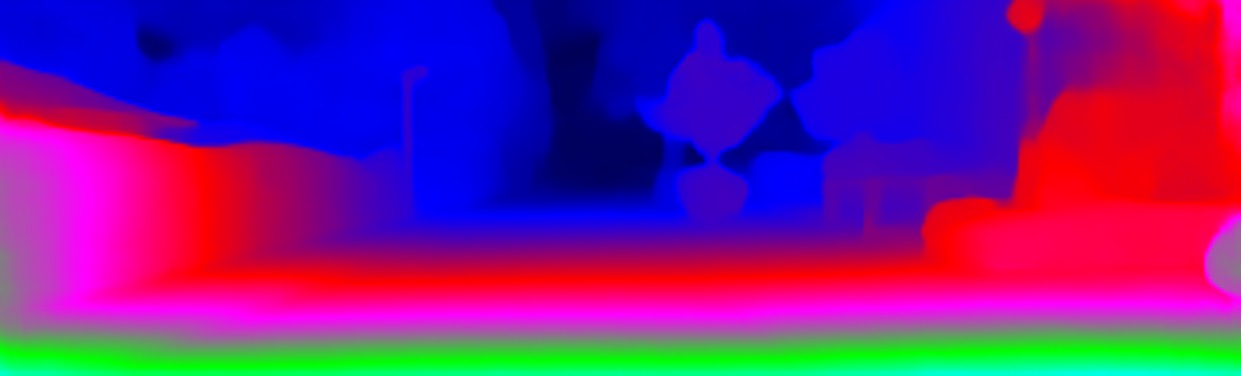} \\ 

        \includegraphics[width=0.24\textwidth]{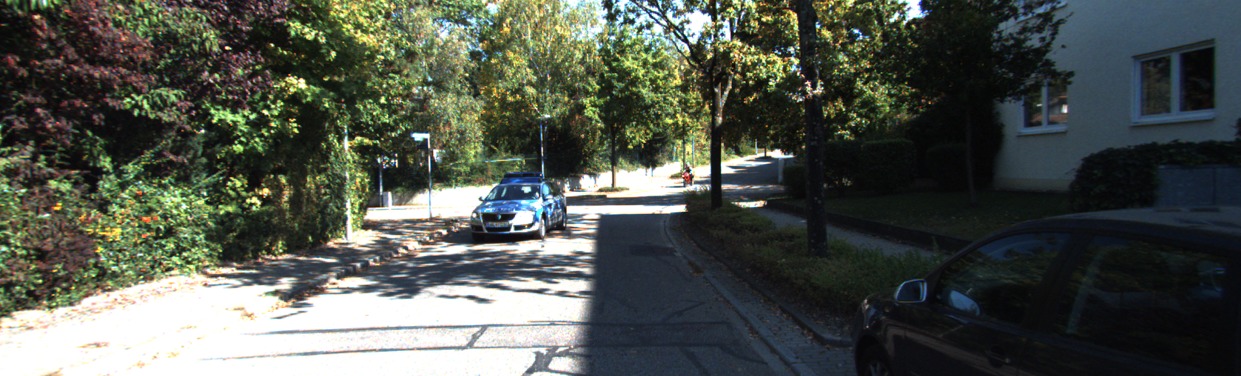} & \includegraphics[width=0.24\textwidth]{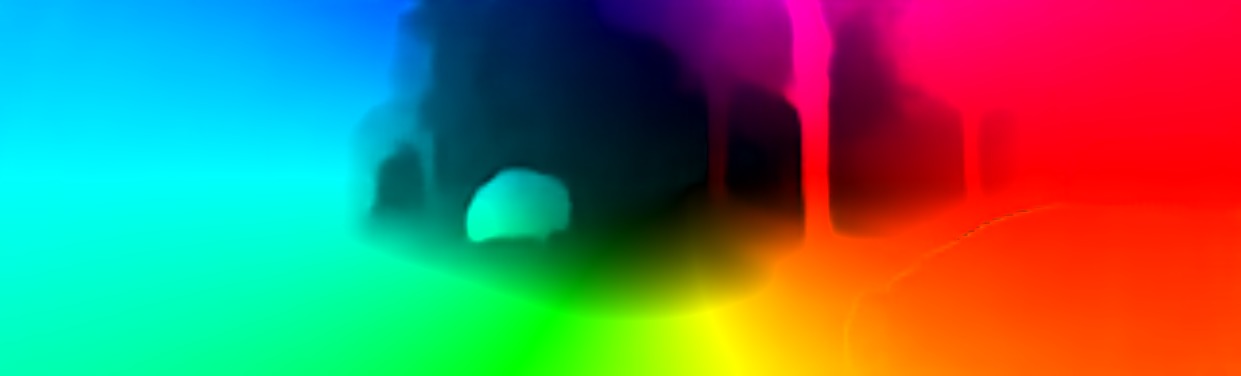} &
        \includegraphics[width=0.24\textwidth]{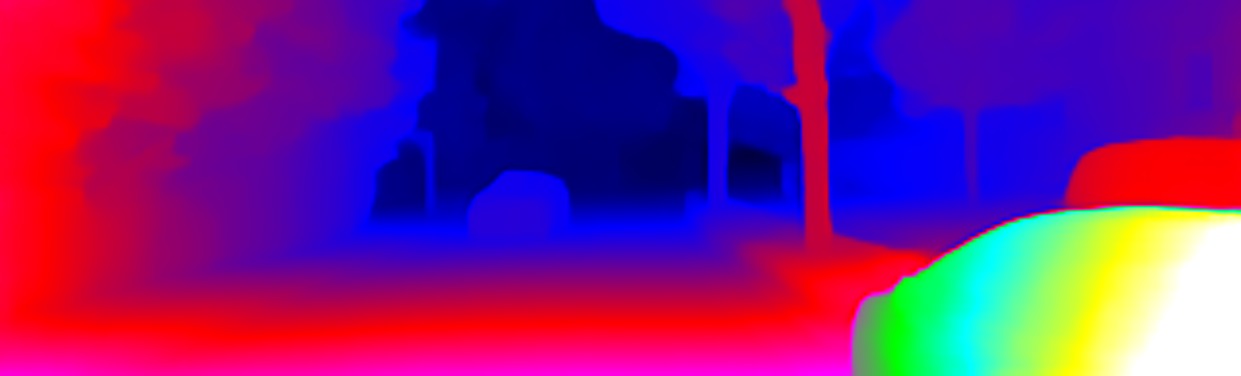} &
        \includegraphics[width=0.24\textwidth]{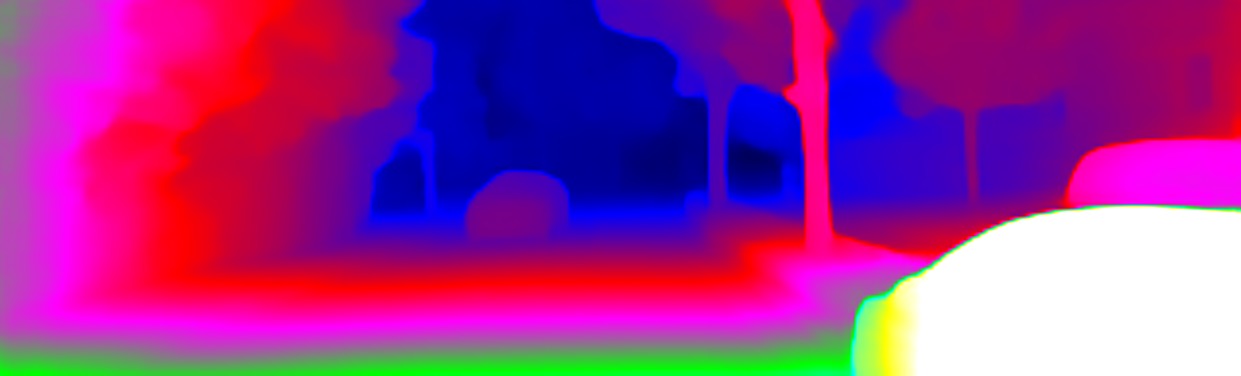} \\         

        \includegraphics[width=0.24\textwidth]{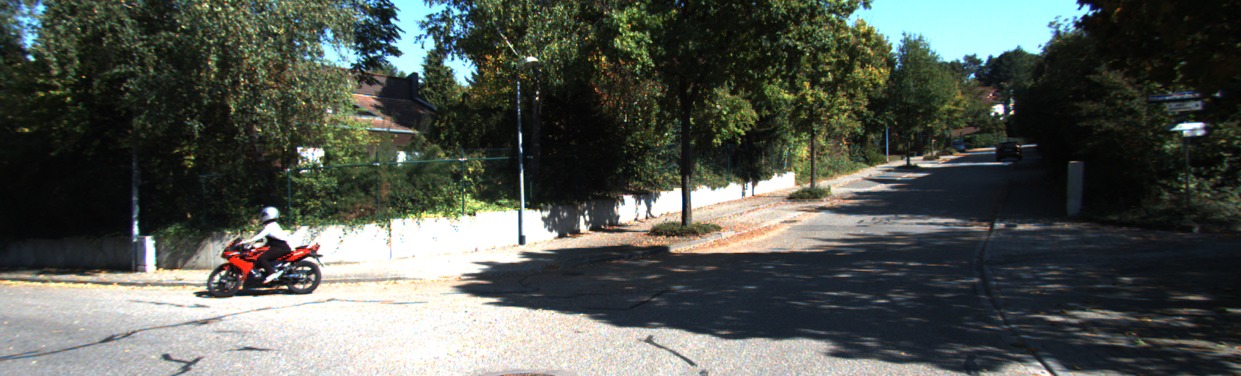} & \includegraphics[width=0.24\textwidth]{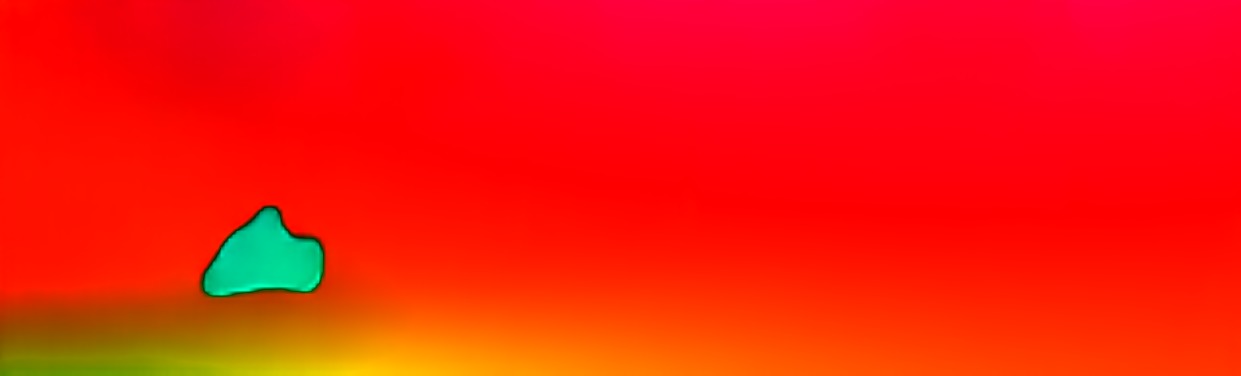} &
        \includegraphics[width=0.24\textwidth]{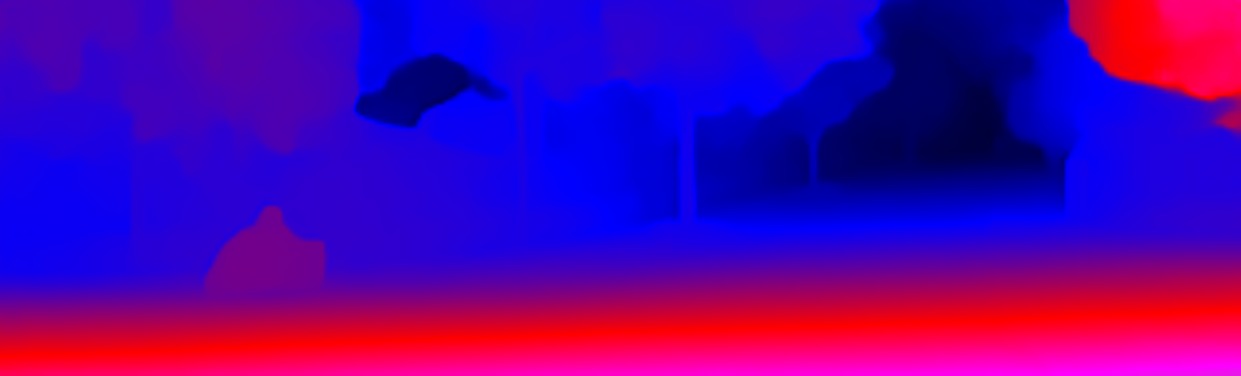} &
        \includegraphics[width=0.24\textwidth]{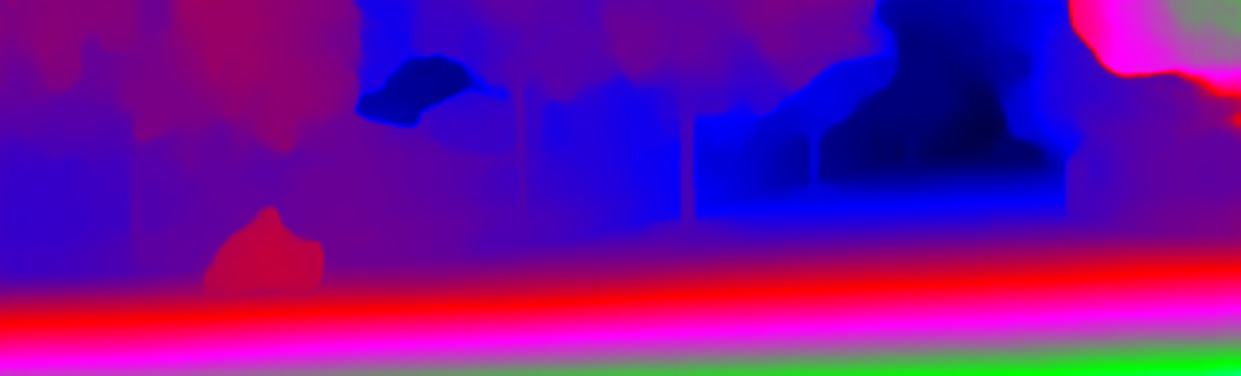} \\         
    \end{tabular}
    \caption{\textbf{Qualitative results on KITTI raw sequence {2011$\_$10$\_$03$\_$drive$\_$0034} \cite{KITTI_RAW}.} From left to right, left image at $t_1$, optical flow, disparity and disparity change.}
    \label{fig:kitti}
\end{figure*}

\section{3D Correlation layer -- limitations}

Although acting as bridge between 2D flow and depth, the 3D correlation layer has some evident limitations. 
The first concerns with computational complexity, since it grows with three search radius. This makes prohibitive to deploy our layer in complex models like U-net networks.
For instance, if we imagine to design an architecture starting from FlowNet \cite{dosovitskiy2015flownet} or DispNet \cite{mayer2016large} to infer full scene flow and to insert a 3D correlation layer, this would introduce an extremely high amount of features to be processed.
Assuming a range of 40 for both flow and disparity at quarter resolution as in the original networks, with stride 2 for the former, leads to 400, 81 and 81 features for flow, disparity and disparity change independently, for a total of 562 features. By simply adding a 3D layer acting on a search range of $40\time40\time1$ and a stride 2, i.e. as range 1 on the third dimension as in our experiments, would increase the features to 962, with prohibitive memory requirements considering the resolution at which they are processed, i.e. quarter. By simply increasing the disparity change search range to 5 ($r_z=2$),  desirable when working at higher resolutions, would add 2000 features instead of 400, up to 2562 total features.
This makes the 3D correlation layer particularly appealing for deployment in pyramidal architectures only.

\section{Additional qualitative results}
We present some additional qualitative results both synthetic and real datasets. Figures \ref{fig:things1} and \ref{fig:things2} depict scene flow estimation on the test set of FlyingThings3D dataset \cite{mayer2016large} after 1.2M steps of training on synthetic images, Figure \ref{fig:kitti} shows results on a sequence taken by the KITTI raw dataset \cite{KITTI_RAW}, i.e. {2011$\_$10$\_$03$\_$drive$\_$0034}, after 50K iterations of fine-tuning on KITTI 2015 training set and finally Figure \ref{fig:weanhall} collects qualitative examples on the WeanHall indoor dataset \cite{weanhall}, using the aforementioned network fine-tuned on KITTI. While results on FlyingThings3D and KITTI show the behavior of DWARF on environments that are similar to those observed during training, we underline that DWARF has never been trained on images similar to those collected in the WeanHall dataset. This experiment confirms that DWARF generalizes quite well to unseen and very different environments.

Finally, we provide an additional \textbf{video sequence} showing DWARF in action on {2011$\_$10$\_$03$\_$drive$\_$0034} and WeanHall sequences from which qualitative examples in Figures \ref{fig:kitti} and \ref{fig:weanhall} have been sampled. The video is available at \UrlFont{\textbf{https://www.youtube.com/watch?v=qGWpi3z2M74}}.

\begin{figure*}[h]
    \centering
    \renewcommand{\tabcolsep}{1pt}
    \begin{tabular}{cccc}

        \includegraphics[width=0.24\textwidth]{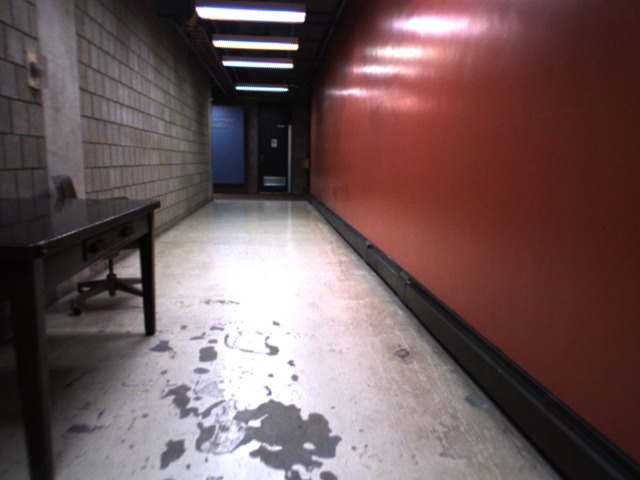} & \includegraphics[width=0.24\textwidth]{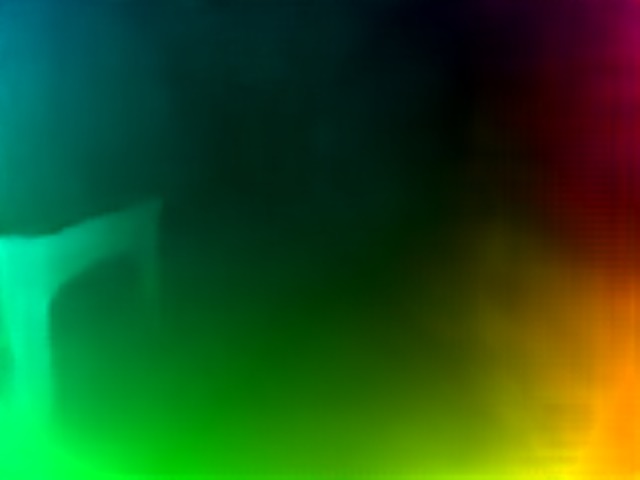} &
        \includegraphics[width=0.24\textwidth]{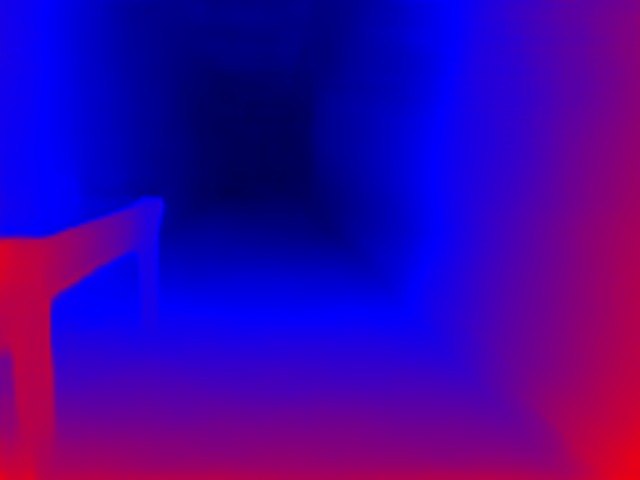} &
        \includegraphics[width=0.24\textwidth]{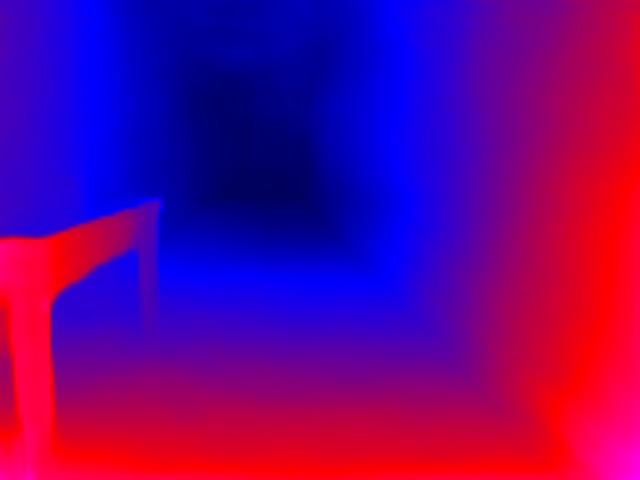} \\ 

        \includegraphics[width=0.24\textwidth]{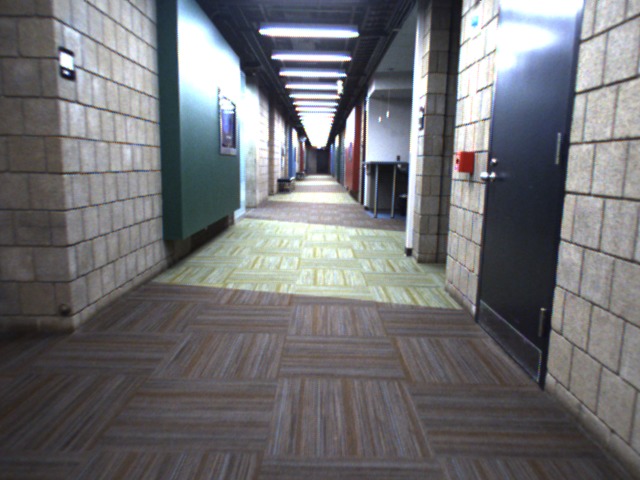} & \includegraphics[width=0.24\textwidth]{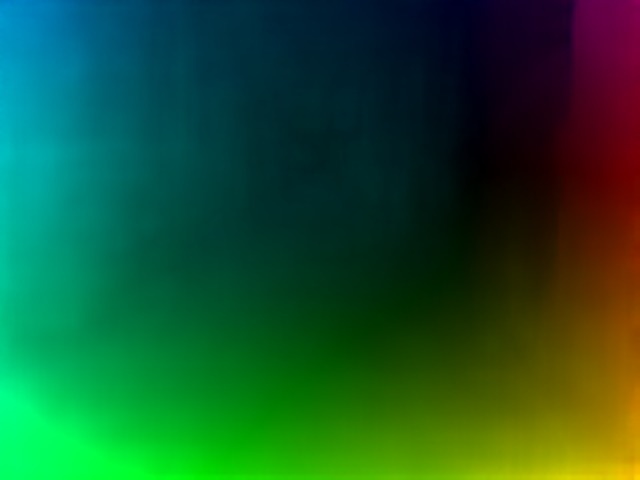} &
        \includegraphics[width=0.24\textwidth]{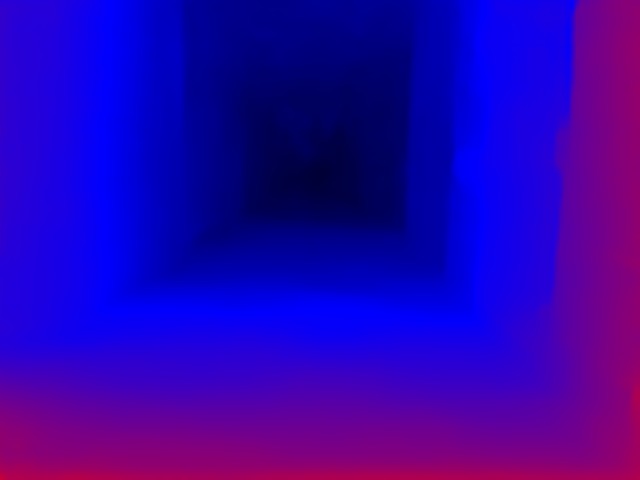} &
        \includegraphics[width=0.24\textwidth]{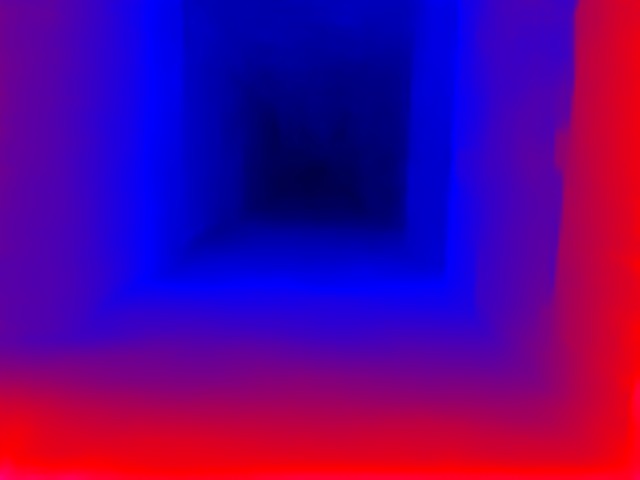} \\ 

        \includegraphics[width=0.24\textwidth]{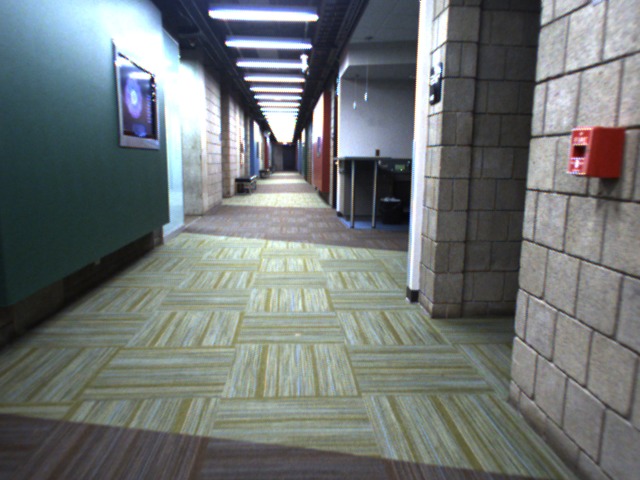} & \includegraphics[width=0.24\textwidth]{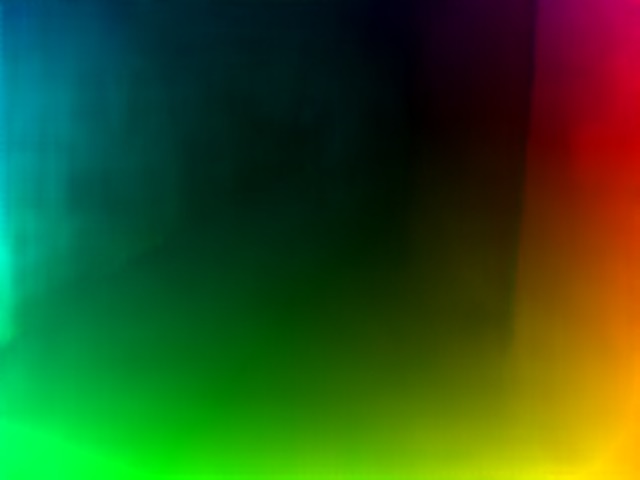} &
        \includegraphics[width=0.24\textwidth]{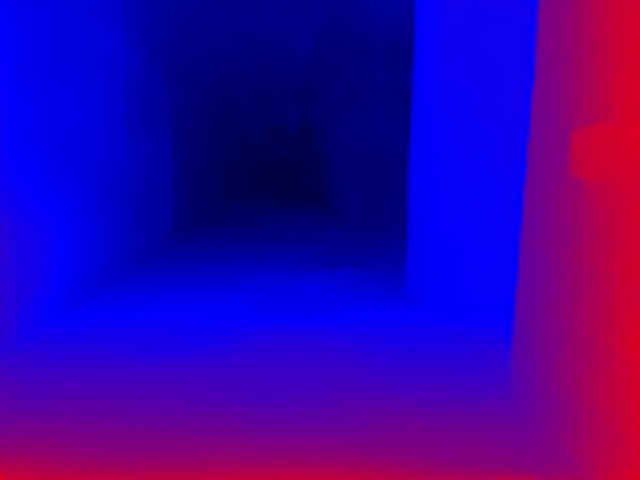} &
        \includegraphics[width=0.24\textwidth]{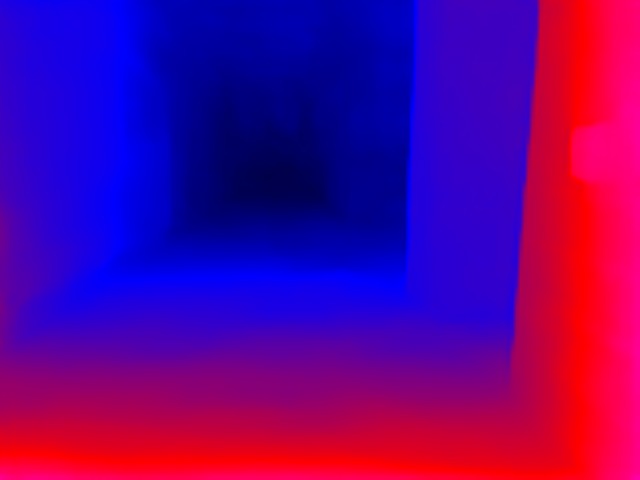} \\ 

        \includegraphics[width=0.24\textwidth]{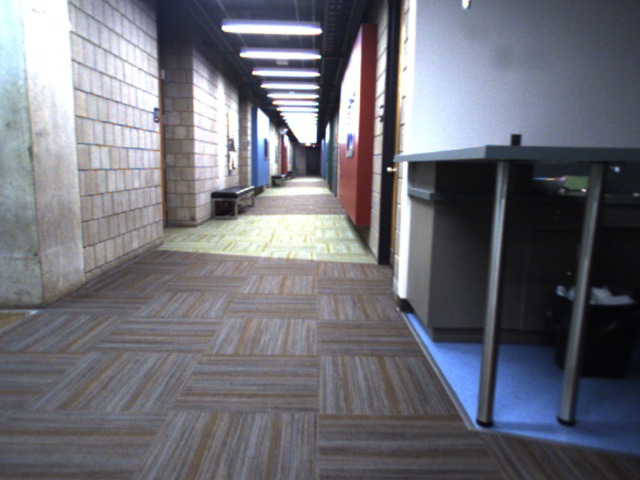} & \includegraphics[width=0.24\textwidth]{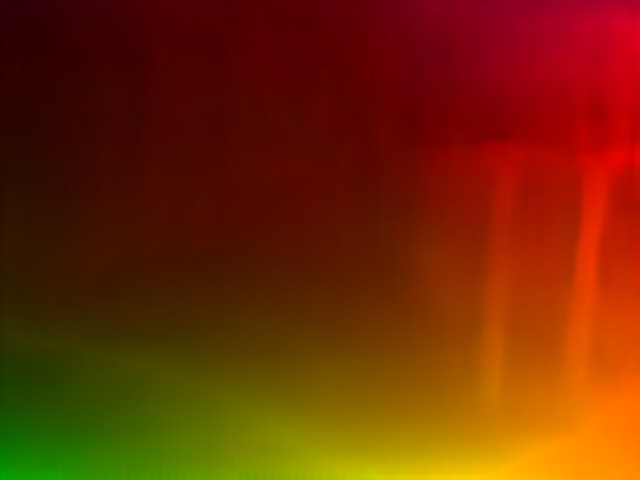} &
        \includegraphics[width=0.24\textwidth]{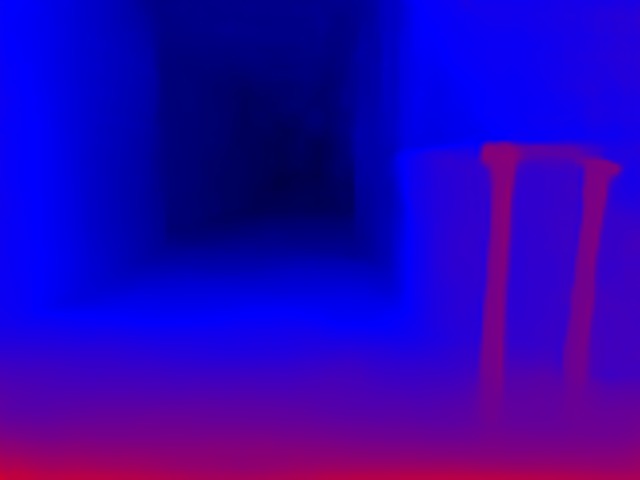} &
        \includegraphics[width=0.24\textwidth]{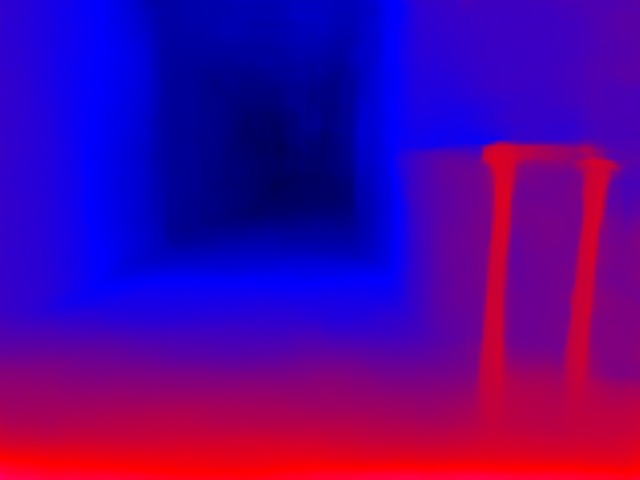} \\ 
    \end{tabular}
    \caption{\textbf{Qualitative results on WeanHall dataset \cite{weanhall}.} From left to right, left image at $t_1$, optical flow, disparity and disparity change.}
    \label{fig:weanhall}
\end{figure*}

\end{document}